\documentclass{article}
\usepackage[numbers,sort&compress]{natbib}
\usepackage{iclr2027_conference,times}
\iclrfinalcopy

\usepackage[utf8]{inputenc} 
\usepackage[T1]{fontenc}    
\usepackage{nicefrac}       
\usepackage{microtype}      
\usepackage{xcolor}         
\usepackage{hyperref}
\usepackage{url}
\usepackage{booktabs}       
\usepackage{amsfonts}       
\usepackage{amsmath}
\usepackage{multirow}    
\usepackage{algorithmic} 
\usepackage{pifont}      
\usepackage{tablefootnote} 
\usepackage{enumitem}
\usepackage{graphicx}
\usepackage{amssymb}  

\usepackage{algorithm}
\usepackage{subcaption}

\newtheorem{theorem}{Theorem}[section]

\title{NoiseRater: Meta-Learned Noise Valuation for Diffusion Model Training} 
\author{
\textbf{Haokai Zhao}$^{1}$,\,
\textbf{Da Xing}$^{2}$,\,
\textbf{Hanqun Cao}$^{3}$,\,
\textbf{Tinson Xu}$^{4}$,\,
\textbf{Xinyu Xiang}$^{5}$ \\
\textbf{Yanchao Li}$^{3}$,\,
\textbf{Xiangru Tang}$^{3}$,\,
\textbf{Hongbin Lin}$^{3}$,\,
\textbf{Zehong Wang}$^{6}$,\,
\textbf{Kuan Pang}$^{5}$ \\
\textbf{Peng Xia}$^{7}$,\,
\textbf{Molei Tao}$^{8}$,\,
\textbf{Li Erran Li}$^{3}$,\,
\textbf{Aditya Joshi}$^{1}$,\,
\textbf{Jure Leskovec}$^{5}$ 
\textbf{Fang Wu}$^{5}$\thanks{Corresponding author.} \\
$^{1}$University of New South Wales,\,
$^{2}$Johns Hopkins University \\
$^{3}$Independent Researcher,\,
$^{4}$University of Chicago,\,
$^{5}$Stanford University \\
$^{6}$University of Notre Dame,\,
$^{7}$UNC--Chapel Hill,\,
$^{8}$Georgia Institute of Technology \\
\texttt{fangwu97@stanford.edu}
}

\begin{document}
\maketitle

\begin{abstract}
Training a diffusion model involves two sources of randomness for each data sample: the timestep and the Gaussian noise realization. The timestep has been studied extensively through scheduling and weighting, whereas the impact of the noise realization at a given timestep is still underexplored. In this work, we examine whether different noise instances are equally informative. We introduce \emph{NoiseRater}, a network that scores an individual noise instance conditioned on the data sample and timestep. The rater is learned through bilevel optimization, where its scores reweight the diffusion loss in the inner loop, and it is updated to reduce validation loss after the inner-loop updates. Using the trained rater to select training noise, we observe three properties of training noise. First, noise realizations at the same timestep are not equally useful: the rater's top-scored noise improves performance over i.i.d.\ sampling, while its bottom-scored noise degrades it. {Second, this utility is contextual: it depends most strongly on the timestep, and to a lesser extent on the image and the class label.} Third, noise selection is complementary to timestep-level design, retaining most of its gain when combined with existing scheduling and weighting schemes. These findings establish instance-level noise valuation as a new axis for understanding and improving diffusion training.
Code is available at~\url{https://github.com/JoeZhao527/Noise-Rater}. 
\end{abstract}
\section{Introduction}

Diffusion models~\citep{sohl2015deep,song2020score,ho2020denoising} have emerged as a dominant paradigm for generative modeling~\citep{rombach2022high,polyak2024movie,watson2023novo}. Recently, noise has increasingly become a central object of control at inference time. Methods focusing on test-time compute scaling~\citep{he2025scaling,ren2025driftlite,kim2025inference} and noise trajectory manipulation~\citep{li2025dynamic,ahn2024noise,singhal2025general} demonstrate that the specific magnitude, structure, and evolution of noise critically shape the final generation quality. In contrast to this fine-grained control at inference time, standard training treats the choice of noise as largely indiscriminate.

During standard training, the diffusion objective relies on two independent sources of stochasticity for any given data sample: the noise level (timestep) and the noise realization itself. While the former has been extensively optimized through timestep scheduling and weighting~\citep{choi2022perception,hang2023efficient,karras2022elucidating,esser2024scaling,hang2025improved,raya2026information,sun2026variance,kim2024denoising}, the latter remains largely overlooked. Once the timestep is fixed, the noise is drawn independently from a standard Gaussian, and every realization contributes equally to the loss. This implicitly assumes that all noise realizations are equally useful for learning, leaving fine-grained \emph{instance-level} noise variation unexplored.

This raises a fundamental question: \textbf{\emph{are all noise realizations equally useful for learning diffusion models?}} More specifically, even at the same timestep, different noise instances may carry varying levels of learning signal. Some may provide clear supervision for denoising, while others may be ambiguous, redundant, or less informative for optimization. If so, uniformly treating noise during training may lead to suboptimal results.

In this work, we explore the problem of instance-level noise valuation during training using meta-learning. Concretely, we introduce \emph{NoiseRater}, a parametric function that assigns a score to each noise instance conditioned on the data sample and timestep. These scores are used to reweight the diffusion loss, allowing the model to focus on more informative noise while downweighting less useful ones. To train the NoiseRater, we formulate a bilevel optimization framework~\citep{maclaurin2015gradient,engstrom2025optimizing}: the diffusion model is trained using weighted noise samples in the inner loop, while the rater is optimized in the outer loop to minimize validation loss. This setup enables the rater to identify the contribution of distinct noise instances to generalization performance, rather than relying on heuristic criteria.

By integrating NoiseRater into the training pipeline, we conduct a comprehensive empirical study of how instance-level noise affects diffusion learning, yielding three key insights about the utility of noise realizations:
1) \emph{Not equally useful.} Even for the same data sample at the same timestep, different Gaussian realizations carry different learning signals. Selecting the rater's top-scored candidate improves over the uniformly random choice made by standard training, and selecting its bottom-scored candidate falls below it at the same per-step cost.
2) \emph{Context-dependent.} The utility of a noise realization is not an intrinsic property of the noise itself. {Rather, it depends most strongly on the timestep, and to a lesser extent on the data sample: proxies computed from the noise alone, such as magnitude-based selection, capture only a fraction of the attainable gain, and most of the rater's advantage is lost when its conditioning on the timestep is removed.}
3) \emph{Complementary.} Instance-level noise selection addresses an axis largely distinct from timestep-level design: combined with established timestep weighting and scheduling schemes, it retains most of its standalone improvement. Our contributions are summarized as follows:

\begin{itemize}
\item We formalize instance-level noise valuation and introduce \emph{NoiseRater}, a meta-learned rater that measures the utility of a noise instance and can be used as a noise selector to improve diffusion training.
\item We show that noise realizations at the same timestep are not equally useful. Selecting the rater's top-scored candidate improves over i.i.d.\ sampling, and selecting its bottom-scored candidate falls below it.
\item {We show that noise utility is context-dependent. It is not a property of the noise alone, but depends most strongly on the timestep, and to a lesser extent on the data sample.}
\item We show that noise selection is complementary to timestep-level design, retaining most of its gain on top of existing weighting and scheduling schemes.
\end{itemize}

\begin{figure}[t]
    \centering
    \includegraphics[width=1.0\linewidth]{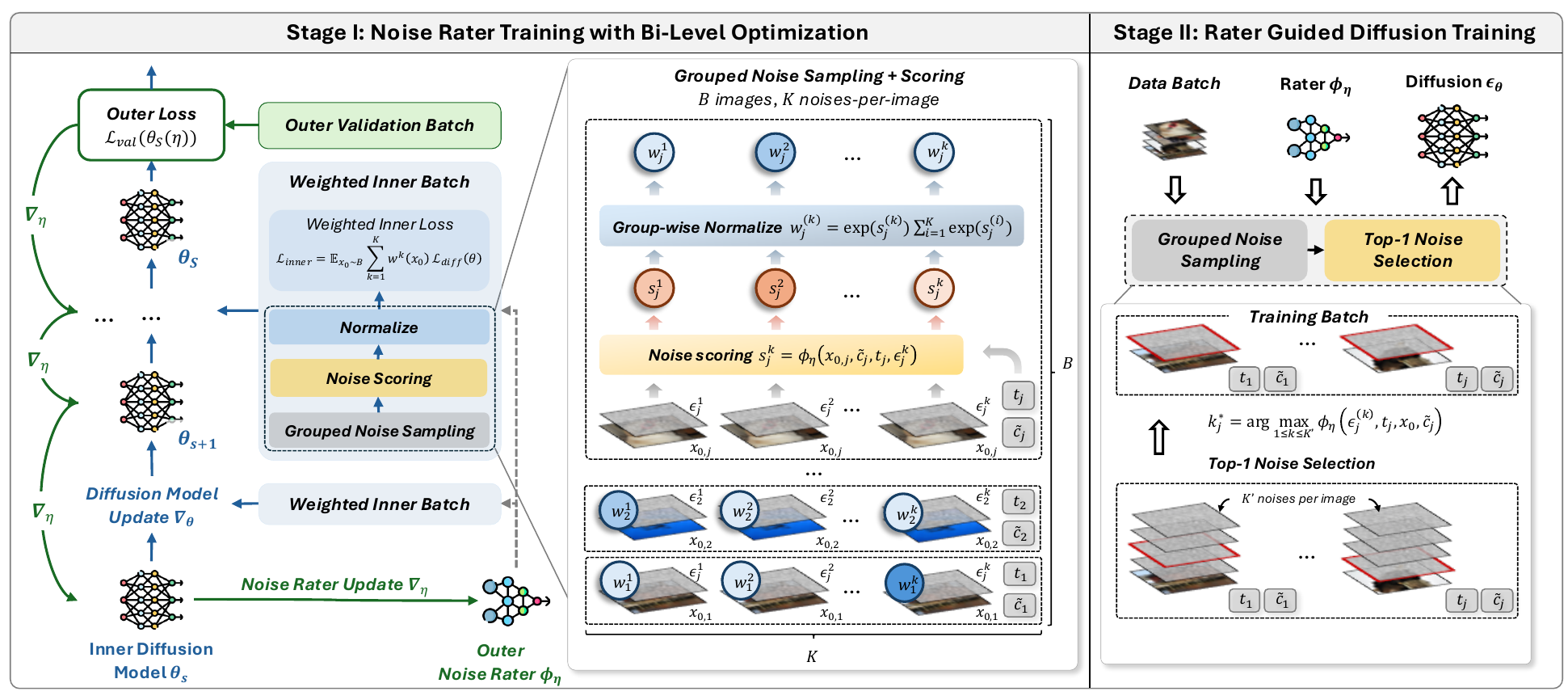}
    \caption{Illustration of our meta-learned noise valuation approach for diffusion training. We first train a noise rater using bilevel optimization, and then guide diffusion training with the rater's scores. }
    \label{fig:main}
\end{figure}

\section{Preliminaries: Diffusion Models}

Diffusion models define a generative process by learning to reverse a gradual corruption of data into noise. In the standard formulation, a forward noising process incrementally perturbs a clean data sample $\mathbf{x}_0 \sim q(\mathbf{x}_0)$ into a sequence of increasingly noisy latent variables $\{\mathbf{x}_t\}_{t=1}^T$.

\paragraph{Forward process.}
The forward diffusion process is a Markov chain that adds Gaussian noise at each timestep $q(\mathbf{x}_t \mid \mathbf{x}_{t-1}) = \mathcal{N}\big(\mathbf{x}_t; \sqrt{1-\beta_t}\, \mathbf{x}_{t-1}, \beta_t \mathbf{I}\big)$, $t = 1, \dots, T$, where $\{\beta_t\}_{t=1}^T$ is a predefined variance schedule controlling the noise magnitude at each step. Intuitively, this process gradually destroys the structure in $\mathbf{x}_0$, eventually producing a nearly isotropic Gaussian distribution.

A key property of this construction is that the marginal distribution at any timestep $t$ admits a closed-form expression $q(\mathbf{x}_t \mid \mathbf{x}_0) = \mathcal{N}\big(\mathbf{x}_t; \sqrt{\bar{\alpha}_t} \mathbf{x}_0, (1-\bar{\alpha}_t)\mathbf{I}\big)$, where $\alpha_t = 1 - \beta_t$ and $\bar{\alpha}_t = \prod_{s=1}^t \alpha_s$. This expression allows us to directly sample $\mathbf{x}_t$ from $\mathbf{x}_0$ without simulating the full Markov chain, which is crucial for efficient training.

\paragraph{Reverse process and training objective.}
The generative model is trained to invert this noising process. Instead of directly parameterizing the reverse transition $p_\theta(\mathbf{x}_{t-1} \mid \mathbf{x}_t)$, it is common to train a neural network $\epsilon_\theta(\mathbf{x}_t, t)$ to predict the noise that was added to produce $\mathbf{x}_t$. Specifically, given a noisy sample $\mathbf{x}_t = \sqrt{\bar{\alpha}_t} \mathbf{x}_0 + \sqrt{1-\bar{\alpha}_t}\, \epsilon$, $\epsilon \sim \mathcal{N}(0, I)$, the model is trained using a simple mean squared error (MSE) objective:
\begin{align}
\label{equ:diff_loss}
\mathcal{L}_{\text{diff}}(\theta) = \mathbb{E}_{t, \mathbf{x}_0, \epsilon} \left[ \left\| \epsilon - \epsilon_\theta(\mathbf{x}_t, t) \right\|^2 \right].
\end{align}
Training proceeds by sampling triples $(\mathbf{x}_0, t, \epsilon)$, constructing $\mathbf{x}_t$ using the closed-form expression above, and minimizing the discrepancy between the true noise $\epsilon$ and the predicted noise $\epsilon_\theta(\mathbf{x}_t, t)$.

\paragraph{Conditional training and classifier-free guidance.}
In conditional diffusion models, the denoising network additionally takes as input a condition $c \sim p(c)$ (e.g., text prompt), and is trained to predict noise as $\epsilon_\theta(\mathbf{x}_t, t, c)$. 
Following classifier-free guidance (CFG)~\citep{ho2022classifier}, we randomly drop the condition during training with probability $p_{\text{drop}}$, replacing $c$ with a null condition $\varnothing$. This yields a unified model that supports both conditional and unconditional predictions as $\mathcal{L}_{\text{diff}}(\theta) = \mathbb{E}_{\mathbf{x}_0, t, \epsilon, c} \left[\|\epsilon - \epsilon_\theta(\mathbf{x}_t, t, \tilde{c})\|^2 \right]$, where $\tilde{c} \in \{c, \varnothing\}$ denotes the possibly dropped condition.

\paragraph{From DDPM to DDIM.}
While the above formulation follows the standard DDPM~\citep{ho2020denoising}, this work implements the DDIM framework~\citep{song2020denoising}. DDIM provides a deterministic (or partially stochastic) non-Markovian sampling procedure that shares the same training objective but enables faster and more flexible generation. Importantly, our modifications operate entirely at the level of the training loss and are therefore directly compatible with DDIM without altering its sampling dynamics.
\section{Method: Meta-Learned Noise Rater}

A central observation about standard diffusion training is that all noise samples are treated equally when optimizing Equ.~\ref{equ:diff_loss}. We hypothesize that not all noise realizations contribute equally to learning: \emph{some may be more informative, while others may be redundant or even detrimental}. To explore this, we introduce a parametric \emph{NoiseRater} $\phi_\eta(\epsilon, t, \mathbf{x}_0, \tilde{c}): \mathcal{E} \times [0,1] \times \mathcal{X} \times \mathcal{C} \rightarrow \mathbb{R}$, which assigns a scalar score to each noise instance conditioned on the clean image $\mathbf{x}_0$, the condition $\tilde{c}$, and the timestep $t$. We adapt meta-learning~\citep{shu2019meta} to train the rater so that its scores reflect the relative importance of the training signal induced by each noise instance.

Our method has two stages (Figure~\ref{fig:main}). In the \emph{rater training} stage, the rater scores each noise instance and these scores reweight the diffusion loss (Section~\ref{sec:noise_weighting}); the rater itself is updated through a bilevel meta-objective defined on the resulting diffusion model (Section~\ref{sec:bilevel_meta_objective}). In the \emph{diffusion training} stage, we freeze the rater, sample a pool of candidate noises for each image--timestep pair, and use the rater's scores to select the best noise (Section~\ref{sec:post_meta_training}).

\subsection{Noise Weighting} \label{sec:noise_weighting}

\paragraph{Grouped Noise Sampling.} To ensure that the noise rater focuses on differences between noise realizations rather than image content, we construct each minibatch $B$ as groups of samples that share the same underlying clean image. Specifically, for each $\mathbf{x}_0$, $\tilde{c}$ and $t$, we draw $K$ independent noises $\{\epsilon^{(k)}\}_{k=1}^K$ (e.g., $K=4$), producing a set of noisy inputs $\{\mathbf{x}_t^{(k)}\}_{k=1}^K$.

We then apply a \emph{group-wise normalization} to these $K$ instances associated with the same $\mathbf{x}_0$. The rater assigns scores $\{\phi_\eta(\epsilon^{(k)}, t, \mathbf{x}_0, \tilde{c})\}_{k=1}^K$, which are normalized via a softmax within the group:
\begin{align}
w^{(k)}(\mathbf{x}_0,\tilde c,t) =  \frac{\exp(\phi_\eta(\epsilon^{(k)}, t, \mathbf{x}_0, \tilde{c}))} {\sum_{k'=1}^K \exp(\phi_\eta(\epsilon^{(k')}, t, \mathbf{x}_0, \tilde{c}))}.
\end{align}
This normalization enforces that weights are assigned based on the relative importance of noise realizations under a fixed condition $(\mathbf{x}_0, \tilde{c}, t)$, preventing the rater from exploiting variations in image content or conditioning across the batch $B$. 

\paragraph{Rater-guided Training Objective.} The standard diffusion loss is modified into a group-wise weighted objective:
\begin{align}
\mathcal{L}_{\text{inner}}(\theta; \eta) = \mathbb{E}_{\mathbf{x}_0 \sim B} \left[ \sum_{k=1}^K  w^{(k)}(\mathbf{x}_0,\tilde c,t) \cdot \|\epsilon^{(k)} - \epsilon_\theta(\mathbf{x}_t^{(k)}, t, \tilde{c})\|^2 \right],
\end{align}
where $\{(\epsilon^{(k)}, \mathbf{x}_t^{(k)})\}_{k=1}^K$ are independently sampled noise instances associated with $(\mathbf{x}_0, \tilde{c}, t)$, and $w^{(k)}(\mathbf{x}_0, \tilde{c}, t)$ are the corresponding group-normalized weights. This formulation reweights training signals according to the relative importance of noise realizations conditioned on $(\mathbf{x}_0, \tilde{c}, t)$, emphasizing informative noise while avoiding spurious correlations with image content or conditioning.

\subsection{Bilevel Meta-Objective}\label{sec:bilevel_meta_objective}

We frame the learning of the noise rater as a bilevel optimization problem~\citep{calian2025datarater}, where the diffusion model $\epsilon_\theta$ and the rater $\phi_\eta$ are optimized at different levels.

\paragraph{Inner optimization} {\texttt{(model training)}.}
For a fixed noise rater $\phi_\eta$, the diffusion model parameters $\theta$ are updated by performing multiple gradient steps on the weighted loss:
\begin{align}
    \theta^{(s+1)} \leftarrow \theta^{(s)} - \alpha \nabla_\theta \mathcal{L}_{\text{inner}}(\theta^{(s)}; \eta), \quad s = 0, \dots, S-1,
\end{align}
where $S\geq 1$ denotes the number of inner-loop updates. 
We write $\theta^{(S)}(\eta)$ for the resulting parameters. 
This corresponds to standard diffusion training, except that each training signal associated with $\mathbf{x}_0$ is reweighted according to the learned importance scores $\{w^{(k)}(\mathbf{x}_0,\tilde c,t)\}_{k=1}^K$.

\paragraph{Outer optimization} {\texttt{(rater training)}.}
The noise rater parameters $\eta$ are updated to improve the downstream performance of the diffusion model after the inner updates:
\begin{align}
    J_S(\eta) = \mathcal{L}_{\text{val}}(\theta^{(S)}(\eta)), \quad \eta^* = \arg\min_\eta J_S(\eta),
\end{align}
where $\mathcal{L}_{\text{val}}$ is computed on held-out data. $J_S$ is what Algorithm~\ref{alg:noise_rater} optimizes; the idealized objective $J(\eta)=\mathcal{L}_{\text{val}}(\theta^*(\eta))$ with $\theta^*=\arg\min_\theta\mathcal{L}_{\text{inner}}$ is analyzed in Thm.~\ref{thm:meta_gradient} and can be viewed as the $S\to\infty$ limit.

Conceptually, the rater $\phi_\eta$ learns to assign higher weights to noise samples that lead to better generalization of the diffusion model $\epsilon_\theta$ after several training updates. This is closely related to meta-learning and data valuation approaches~\citep{shu2019meta,calian2025datarater}, but here the optimization operates over \emph{noise realizations} rather than data points.



\paragraph{Meta-Gradient Computation.} Optimizing $J_S$ requires differentiating through the inner optimization process. By the chain rule,
\begin{align}
\nabla_\eta J_S(\eta)
= \frac{\partial \mathcal{L}_{\text{val}}}{\partial \theta}\Big|_{\theta^{(S)}(\eta)} \cdot \frac{\partial \theta^{(S)}(\eta)}{\partial \eta},
\end{align}
where $\frac{\partial \theta^{(S)}(\eta)}{\partial \eta}$ captures how changes in the rater affect the inner diffusion model after $S$ updates.

We compute it by differentiating through all $S$ gradient steps using mixed-mode differentiation (Mixflow-MG~\citep{kemaev2025scalable}), which keeps the memory cost of the unrolled computation tractable.

\subsection{Post Meta-Training: Rater-Guided Noise Selection}\label{sec:post_meta_training}

After the meta-learning stage, we obtain a trained noise rater $\phi_\eta$ that captures the relative utility of noise instances for improving generalization. We then transition to a standard diffusion training phase in which the rater is \emph{frozen}.

\paragraph{Top-1 noise selection.}
For each data sample $(\mathbf{x}_0, \tilde{c}, t)$, we sample a pool of $K'$ candidate noise instances $\{\epsilon^{(k)}\}_{k=1}^{K'}$ and select the highest-scoring one:
\begin{align}
k^* = {\arg\max}_{1\leq k\leq K'} \; \phi_\eta(\epsilon^{(k)}, t, \mathbf{x}_0, \tilde{c}).
\end{align}
Only $\epsilon^{(k^*)}$ is passed to the diffusion model, so each training step costs one diffusion forward/backward pass plus $K'$ rater evaluations. Selection depends only on the ordering of the raw scores; the group-wise softmax of Section~\ref{sec:noise_weighting} is used solely during meta-training.

\paragraph{Soft weighting as a surrogate for hard selection.}
The soft weighting of Section~\ref{sec:noise_weighting} is a differentiable surrogate for learning a ranking over candidate noises; Top-1 selection is a deployment policy that exploits this ranking. The two are not equivalent, and we validate hard selection empirically (Section~\ref{sec:results}). Note also that the selected noise no longer follows $\mathcal{N}(0, I)$ but a rater-induced distribution $q_\eta(\epsilon \mid \mathbf{x}_0, t, \tilde{c})$, so Stage II minimizes
\begin{align}
\mathbb{E}_{\epsilon \sim q_\eta(\cdot \mid \mathbf{x}_0, t, \tilde{c})}\big[\|\epsilon - \epsilon_\theta(\mathbf{x}_t, t, \tilde{c})\|^2\big]
\label{eq:stage2_objective}
\end{align}
rather than Equ.~\ref{equ:diff_loss}. This shift is intentional and analogous to non-uniform timestep sampling: it biases the training signal toward more useful noise realizations, while the forward corruption and the sampling procedure remain unchanged. {Appx.~\ref{app:denoising_error} bounds this shift and measures its effect on the standard objective.}




\begin{algorithm}[t]
\caption{Meta-Learning a Noise Rater for Diffusion Models}
\label{alg:noise_rater}
\begin{algorithmic}[1]
\STATE \textbf{Inputs:} Training dataset $\mathcal{D}_{\text{train}}$ and validation dataset $\mathcal{D}_{\text{val}}$; initialize noise rater parameters $\eta$ (meta-optimizer $M$) and diffusion model parameters $\theta$ (optimizer $D$); number of outer steps $N$ and inner steps $S$; batch size $B$; group size $K$.
\FOR{outer step $n = 1, \dots, N$} 
    \FOR{inner step $s = 1, \dots, S$}
        \STATE Sample a grouped batch $\mathcal{B} = \{(\mathbf{x}_{0,j}, \tilde{c}_j, t_j, \{\epsilon^{(k)}_j\}_{k=1}^K)\}_{j=1}^B$ from $\mathcal{D}_{\text{train}}$
        
        \STATE For group $j$, evaluate the noise rater scores $s_j^{(k)} = \phi_\eta(\epsilon^{(k)}_j, t_j, \mathbf{x}_{0,j}, \tilde{c}_j)$ for $k=1,\dots,K$
        
        \STATE Convert scores into normalized weights $w_j^{(k)} = \exp(s_j^{(k)}) / \sum_{i=1}^K \exp(s_j^{(i)})$
        \STATE Construct noisy inputs $\mathbf{x}_{t,j}^{(k)}$, and compute the weighted loss
        \STATE \hspace{1em} $\mathcal{L}_\mathcal{B}(\theta) = \mathbb{E}_{j}\big[\sum_{k=1}^K w_j^{(k)} \|\epsilon_j^{(k)} - \epsilon_\theta(\mathbf{x}_{t,j}^{(k)}, t_j, \tilde{c}_j)\|^2\big]$
        
        \STATE Update model parameters $\theta_s \leftarrow \theta_{s-1} - \alpha \nabla_\theta \mathcal{L}_\mathcal{B}(\theta)$ using optimizer $D$
    \ENDFOR
    
    \STATE Sample a validation batch $\mathcal{B}_{\text{val}} \sim \mathcal{D}_{\text{val}}$
    \STATE Compute validation loss $\mathcal{L}_{\text{val}}(\theta_S)$ and update $\eta \leftarrow \eta - \beta \nabla_\eta \mathcal{L}_{\text{val}}(\theta_S)$ using meta-optimizer $M$
\ENDFOR
\RETURN $\eta$
\end{algorithmic}
\end{algorithm}

\begin{theorem}[Meta-gradient for noise-weighted diffusion training]
\label{thm:meta_gradient}
Let $\mathcal{L}_{\mathrm{inner}}(\theta;\eta)$ be the inner objective and $\ell^{(k)}(\theta) = \|\epsilon^{(k)} - \epsilon_\theta(\mathbf{x}_t, t, \tilde{c})\|^2$ be the diffusion loss for the $k$-th noise instance. Assume that $\ell^{(k)}(\theta)$ is twice continuously differentiable in $\theta$, and that $\mathcal{L}_{\mathrm{inner}}(\theta;\eta)$ is $\mu$-strongly convex in $\theta$ for some $\mu > 0$. Then the optimal inner solution $\theta^*(\eta)$ is locally unique and differentiable, and the outer objective $J(\eta) = \mathcal{L}_{\mathrm{val}}(\theta^*(\eta))$ has gradient
\begin{align}
\nabla_\eta J(\eta) = -\Bigl(\nabla_\eta \nabla_\theta \mathcal{L}_{\mathrm{inner}}(\theta^*(\eta);\eta)\Bigr)^\top \Bigl(\nabla_\theta^2 \mathcal{L}_{\mathrm{inner}}(\theta^*(\eta);\eta)\Bigr)^{-1} \nabla_\theta \mathcal{L}_{\mathrm{val}}(\theta^*(\eta)).
\end{align}
\end{theorem}

\paragraph{Interpretation.} Thm.~\ref{thm:meta_gradient} characterizes an idealized bilevel problem in which the inner objective admits a locally unique optimum. These assumptions provide an interpretable implicit-gradient expression but do not generally hold for deep diffusion models. Our implementation instead optimizes a finite-step unrolled meta-objective and differentiates directly through the inner optimization trajectory, without explicitly forming or inverting the inner Hessian.

\begin{theorem}[Gradient alignment principle for noise weighting]
\label{thm:alignment}
Consider a single inner update with learning rate $\alpha > 0$:
$\theta^+(\eta) = \theta - \alpha \sum_{k=1}^K w^{(k)}(\eta)\, \nabla_\theta \ell^{(k)}(\theta)$,
where $\ell^{(k)}(\theta)$ denotes the diffusion loss for the $k$-th noise instance.

Assume that the validation loss $\mathcal{L}_{\mathrm{val}}(\theta)$ has $L$-Lipschitz continuous gradients. Then
\begin{align}
\mathcal{L}_{\mathrm{val}}(\theta^+(\eta)) \leq \mathcal{L}_{\mathrm{val}}(\theta) - \alpha \sum_{k=1}^K w^{(k)}(\eta) \left\langle \nabla_\theta \mathcal{L}_{\mathrm{val}}(\theta), \nabla_\theta \ell^{(k)}(\theta) \right\rangle + \mathcal{O}(\alpha^2).
\end{align}

Consequently, up to second-order terms, minimizing the post-update validation loss encourages assigning larger weights to noise instances $\epsilon^{(k)}$ whose induced training gradients are better aligned with the validation gradient.
\end{theorem}

\paragraph{Interpretation.}
Thm.~\ref{thm:alignment} reveals that noise weighting follows a \emph{gradient alignment principle}. Each noise instance $\epsilon^{(k)}$ contributes in proportion to $\left\langle \nabla_\theta \mathcal{L}_{\mathrm{val}}(\theta), \nabla_\theta \ell^{(k)}(\theta) \right\rangle,$ which measures how well its induced training gradient aligns with the validation objective.
Minimizing the post-update validation loss, therefore, encourages assigning larger weights to noise instances whose gradients are better aligned with improving generalization, while downweighting conflicting or uninformative ones. In effect, the noise rater learns a directionally optimized gradient estimator that amplifies useful learning signals and suppresses detrimental noise.

\section{Results}\label{sec:results}

Throughout the experiments, we observe three properties of training noise: noise realizations are not equally useful (Section~\ref{sec:experiment_unequal}), {their utility depends most strongly on the timestep, and to a lesser extent on the data sample} (Section~\ref{sec:experiment_ctx_dependent}), and noise selection is complementary to timestep-level design (Section~\ref{sec:experiment_complement}). Section~\ref{sec:experiment_improve} {examines the compute cost of noise selection: the gain survives once the per-step scoring cost is included, and Appx.~\ref{app:meta_cost} reports the comparison that also includes the one-time rater training.}

\subsection{Experiment Setups}
\label{sec:experiment_setup}
\paragraph{Datasets and tasks.} We evaluate on two datasets: \textsc{FFHQ}~\citep{karras2019style} at $256{\times}256$ for unconditional generation, and \textsc{ImageNet}~\citep{deng2009imagenet} at $256{\times}256$ for class-conditional generation. {On ImageNet, both the diffusion model and the rater receive the class label, and we report two sampling settings: without CFG (guidance scale $1.0$) and with CFG~\citep{ho2022classifier} at scale $1.25$.}

\paragraph{Models.}
The diffusion model is DiT~\citep{peebles2023scalable} in the latent space of a pre-trained VAE~\citep{rombach2022high}; we use DiT-S/2 for the main results and DiT-B/2 and DiT-L/2 in Appx.~\ref{app:larger_backbone_generalization}. The rater is a dual-stream DiT-S/2 in the style of SD3~\citep{esser2024scaling}: the image latent and the noise are patchified into two token streams that share joint attention, the timestep and class are injected through AdaLN, and the noise-stream tokens are mean-pooled and projected to a scalar score (Appx.~\ref{app:rater_arch_details}, Fig.~\ref{fig:rater_arch}).

\paragraph{Training protocol and evaluation.}
(i) We train DiT-S/2 with standard i.i.d.\ Gaussian noise to a base checkpoint ($400$k steps for ImageNet, $200$k for FFHQ); (ii) we train the rater for $2$k steps on top of this frozen checkpoint, with a $10\%$ random split of the training set as its outer-loop validation set; (iii) we resume diffusion training with rater-selected noise ($k{=}4$) for $80$k steps on ImageNet, reporting FID-50K~\citep{heusel2017gans} every $20$k steps, and for $40$k steps on FFHQ, reporting every $10$k. Hyperparameters are in Appx.~\ref{app:hyperparameters}.

\subsection{Noise realizations are not equally useful}
\label{sec:experiment_unequal}
Table~\ref{tab:strategy_comparison} compares five noise selection strategies. Vanilla is standard training with a single i.i.d.\ noise; the other four draw $k=4$ candidates per sample and keep one. Naive (min) and Naive (max) keep the candidate with the smallest and largest norm, and Rater (bottom) and Rater (top) keep the candidate with the lowest and highest \emph{NoiseRater} score. At every checkpoint, Rater (top) improves over Vanilla and Rater (bottom) falls below it, which shows that noise realizations at the same timestep are not equally useful and that the rater's score orders them by utility. Norm carries some signal, as Naive (max) helps and Naive (min) hurts, but the rater's gain over Vanilla is roughly twice that of Naive (max) on ImageNet. {These gains are repeatable across independent runs (Appx.~\ref{app:seeds}). We further report and discuss the coverage metrics in Appx.~\ref{app:diversity}.}

\begin{table*}[t]
  \centering
  \small
  \caption{Comparison of noise-selection strategies for DiT-S/2 on FFHQ and ImageNet ($256{\times}256$). FID-50K reported along training; lower is better.}
  \label{tab:strategy_comparison}
  \resizebox{\textwidth}{!}{
  \begin{tabular}{ll ccccc}
      \toprule
      \textbf{Dataset} & \textbf{Step} & \textbf{Vanilla} & \textbf{Naive (min)} & \textbf{Naive (max)} & \textbf{Rater (bottom)} & \textbf{Rater (top)} \\
      \midrule
        \multirow{4}{*}{{ImageNet, w/o CFG}}
          & +20k & 63.60 & 64.69 & 62.65 & 65.66 & \textbf{61.88} \\
          & +40k & 62.93 & 64.50 & 62.00 & 65.51 & \textbf{61.03} \\
          & +60k & 62.26 & 63.97 & 61.41 & 64.85 & \textbf{60.37} \\
          & +80k & 61.69 & 63.31 & 60.82 & 64.43 & \textbf{59.97} \\
      \midrule
        \multirow{4}{*}{{ImageNet, w/ CFG}}
          & +20k & 52.02 & 53.28 & 51.29 & 53.96 & \textbf{50.71} \\
          & +40k & 51.31 & 53.02 & 50.47 & 53.70 & \textbf{49.88} \\
          & +60k & 50.61 & 52.52 & 49.95 & 53.12 & \textbf{49.11} \\
          & +80k & 50.12 & 51.82 & 49.29 & 52.67 & \textbf{48.65} \\
      \midrule
      \multirow{4}{*}{FFHQ, Uncond.}
          & +10k & 11.98 & 13.14 & \textbf{11.47} & 13.22 & 11.54 \\
          & +20k & 11.72 & 13.54 & 11.32 & 13.92 & \textbf{11.28} \\
          & +30k & 11.57 & 13.66 & 11.18 & 14.36 & \textbf{11.09} \\
          & +40k & 11.36 & 13.62 & 11.04 & 14.55 & \textbf{10.93} \\
      \bottomrule
  \end{tabular}}
\end{table*}

\subsection{{Noise utility depends on the timestep and the data sample}}
\label{sec:experiment_ctx_dependent}
To test which context the rater relies on, we ablate its inputs and retrain it (1) without the timestep embedding, {(2) without the class label, (3) without the image, (4) without both the image and the class label, and (5) with the noise alone} (Table~\ref{tab:rater_arch_ablation}). Removing the timestep causes the largest degradation, from 59.97 to 61.40 at +80k, which matches the noise-only rater: the image and class label help only when the timestep is present. The noise-only rater retains a small gain over Vanilla (61.40 vs.\ 61.69), so what the rater learns beyond magnitude is almost entirely contextual. {The rater without the class label retains $1.59$ of the $1.72$ FID gain over Vanilla at +80k.} {Together, these results show that noise utility depends most strongly on the timestep, and to a lesser extent on the image and class label, rather than on the noise alone.}

\begin{table}[t]
    \centering
    \small
    \caption{Ablation of rater's input conditions. We progressively remove the timestep embedding, image input, and class label. FID-50K is reported every 20k steps. All raters are trained for 2k steps.}
    \label{tab:rater_arch_ablation}
    \resizebox{0.9\textwidth}{!}{
    \begin{tabular}{lccc cccc}
        \toprule
        & & & & \multicolumn{4}{c}{ImageNet $256{\times}256$, FID-50K $\downarrow$} \\
        \cmidrule(lr){5-8}
        Variant & Timestep & Image & Class & +20k & +40k & +60k & +80k \\
        \midrule
        Full (ours)                  & \ding{51} & \ding{51} & \ding{51} & \textbf{61.88} & \textbf{61.03} & \textbf{60.37} & \textbf{59.97} \\
        w/o timestep                 & \ding{55} & \ding{51} & \ding{51} & 62.73          & 62.22          & 61.74          & 61.40          \\
        {w/o class}         & {\ding{51}} & {\ding{51}} & {\ding{55}} & {\textbf{61.88}} & {61.25} & {60.72} & {60.10} \\
        w/o image                    & \ding{51} & \ding{55} & \ding{51} & 62.08          & 61.28          & 60.66          & 60.29          \\
        w/o image \& class           & \ding{51} & \ding{55} & \ding{55} & 62.05          & 61.43          & 60.81          & 60.39          \\
        w/o image, timestep \& class & \ding{55} & \ding{55} & \ding{55} & 62.74          & 62.33          & 61.94          & 61.40          \\
        \bottomrule
    \end{tabular}}
\vspace{-1em}
\end{table}

\subsection{Noise selection is complementary to timestep-level design}
\label{sec:experiment_complement}
Table~\ref{tab:timestep_composition} combines NoiseRater (top-1 selection from $k=4$ candidates) with three established timestep-level designs: Min-SNR loss weighting \citep{hang2023efficient}, P2 loss weighting \citep{choi2022perception}, and the logit-normal timestep sampling of SD3 \citep{esser2024scaling}. Each scheme improves over vanilla training on its own, and adding NoiseRater improves each of them further at every checkpoint. The gain from noise selection is largely preserved across schemes: its FID reduction at +80k goes from $-1.72$ on vanilla training to between $-1.36$ and $-1.51$ on the three schemes, i.e., roughly 80--90\% of the standalone gain is retained even on top of the strongest scheme. This indicates that instance-level noise selection addresses an axis of the training objective that timestep weighting and scheduling do not reach. {Appx.~\ref{app:seeds} repeats this comparison over independent runs with the $p{=}4$ rater to validate the stability of the gain.}

\begin{table}[t]
  \centering
  \small
  \caption{Noise selection is complementary to timestep-level design.
  FID-50K for DiT-S/2 on ImageNet ($256{\times}256$, {class-conditional without CFG}); each
  pair differs only in adding NoiseRater top-1 selection ($K{=}4$) on top of
  the timestep-level scheme. $\Delta$ is the FID reduction from adding
  NoiseRater at +80k.}
  \label{tab:timestep_composition}
  \begin{tabular}{l cccc c}
      \toprule
      & \multicolumn{4}{c}{ImageNet $256{\times}256$, FID-50K $\downarrow$} & \\
      \cmidrule(lr){2-5}
      \textbf{Method} & \textbf{+20k} & \textbf{+40k} & \textbf{+60k} & \textbf{+80k} & $\Delta_{+80k}$ \\
      \midrule
      Vanilla                & 63.60 & 62.93 & 62.26 & 61.69 & \multirow{2}{*}{$-1.72$} \\
      \;\;+ NoiseRater       & \textbf{61.88} & \textbf{61.03} & \textbf{60.37} & \textbf{59.97} & \\
      \midrule
      Min-SNR                & 62.95 & 61.41 & 60.18 & 59.05 & \multirow{2}{*}{$-1.51$} \\
      \;\;+ NoiseRater       & \textbf{61.17} & \textbf{59.60} & \textbf{58.42} & \textbf{57.54} & \\
      \midrule
      P2                     & 62.37 & 60.20 & 58.72 & 57.45 & \multirow{2}{*}{$-1.40$} \\
      \;\;+ NoiseRater       & \textbf{60.51} & \textbf{58.46} & \textbf{57.06} & \textbf{56.05} & \\
      \midrule
      Logit-normal           & 61.75 & 59.72 & 58.14 & 57.16 & \multirow{2}{*}{$-1.36$} \\
      \;\;+ NoiseRater       & \textbf{60.38} & \textbf{58.29} & \textbf{56.77} & \textbf{55.80} & \\
      \bottomrule
  \end{tabular}
\end{table}

\subsection{The compute cost of noise selection.}
\label{sec:experiment_improve}
The previous sections establish that \emph{NoiseRater} selects noise of higher utility. However, deploying \emph{NoiseRater} adds compute, including a one-time cost for rater training and, at every diffusion step, $k$ rater evaluations for candidate scoring. This section examines the resulting trade-off along two axes, the candidate pool size $k$ and the rater size, and reports each configuration under both equal training steps and equal compute. Equal-step results isolate the quality of the rater's noise valuation; equal-compute results ask whether that valuation still pays off once scoring overhead is charged. Equal compute counts the per-step scoring cost but not the one-time rater training ($\sim$16 GPU-h, Appx.~\ref{app:compute_resource}), which is amortized across runs. {Appx.~\ref{app:meta_cost} reports the comparison with the cost of rater training included, and Appx.~\ref{app:larger_backbone_generalization} tests the transfer of a rater to larger backbones.}

\paragraph{Effect of $k$.}
Table~\ref{tab:effect_of_k_compute} compares candidate pool sizes. A larger pool raises the chance of containing a high-utility noise but costs $k$ rater evaluations per step. At equal steps, $k=4$ is best throughout, and $k=8$ is worse than $k=4$ from +40k onward despite the larger pool. We hypothesize that top-1 selection from a large pool narrows the training noise distribution $q_\eta$ (Eq.~\ref{eq:stage2_objective}) too far from $\mathcal{N}(0,I)$, trading diversity of the training signal for per-sample utility. At equal compute, $k=4$ remains the best trade-off, and $k=8$ barely improves after 6.7 GPU-h.

\begin{table}[t]
\centering
\caption{Effect of the candidate pool size $k$ under equal steps and equal compute, for DiT-S/2 on ImageNet $256{\times}256$ resuming from the 400k
checkpoint.}
\label{tab:effect_of_k_compute}
\resizebox{0.75\linewidth}{!}{
\begin{tabular}{l l cccc}
    \toprule
    & & \multicolumn{4}{c}{ImageNet $256{\times}256$, FID-50K $\downarrow$} \\
    \midrule
    \multirow{5}{*}{\shortstack[l]{Equal\\steps}}
    & Strategy           & +20k & +40k & +60k & +80k \\
    \cmidrule(lr){2-6}
    & $k{=}2$            & 62.50 & 61.68 & 61.07 & 60.38 \\
    & $k{=}4$            & 61.88 & \textbf{61.03} & \textbf{60.37} & \textbf{59.97} \\
    & $k{=}8$            & \textbf{61.76} & 61.74 & 61.41 & 60.97 \\
    \cmidrule(lr){2-6}
    & No Rater ($k{=}1$) & 63.60 & 62.93 & 62.26 & 61.69 \\
    \midrule
    \multirow{5}{*}{\shortstack[l]{Equal\\compute}}
    & Strategy           & 6.7 GPU-h & 9.0 GPU-h & 13.5 GPU-h & 17.6 GPU-h \\
    \cmidrule(lr){2-6}
    & $k{=}2$            & 62.35 & 62.01 & 61.44 & 60.98 \\
    & $k{=}4$            & 61.88 & \textbf{61.59} & \textbf{61.03} & \textbf{60.62} \\
    & $k{=}8$            & \textbf{61.86} & 61.76 & 61.75 & 61.74 \\
    \cmidrule(lr){2-6}
    & No Rater ($k{=}1$) & 63.25 & 62.90 & 62.23 & 61.69 \\
    \bottomrule
\end{tabular}}
\vspace{-1em}
\end{table}

\begin{table}[t]
    \centering
    \caption{Effect of the rater's size at equal training steps and compute. $p\uparrow$ indicates patch size, the larger the faster. $h\downarrow$ indicates hidden dimension, the smaller the faster. $d\downarrow$ indicates the transformer depth for the rater, the smaller the faster.}
    \label{tab:effect_of_rater_size_compute}
    \resizebox{0.75\linewidth}{!}{
    \begin{tabular}{l ccc cccc}
        \toprule
        & $p\uparrow$ & $h\downarrow$ & $d\downarrow$ & \multicolumn{4}{c}{ImageNet $256{\times}256$, FID-50K $\downarrow$} \\
        \midrule
        \multirow{6}{*}{\shortstack[l]{Equal\\steps}}
        & & & & +20k & +40k & +60k & +80k \\
        \cmidrule(lr){5-8}
        & 2 & 384 & 12 & \textbf{61.88} & \textbf{61.03} & \textbf{60.37} & \textbf{59.97} \\
        & 4 & 384 & 12 & 62.10 & 61.40 & 60.81 & 60.21 \\
        & 2 & 192 & 6  & 61.97 & 61.50 & 61.02 & 60.64 \\
        & 4 & 192 & 6  & 62.40 & 61.74 & 61.12 & 60.62 \\
        \cmidrule(lr){2-8}
        & \multicolumn{3}{c}{No rater} & 63.60 & 62.93 & 62.26 & 61.69 \\
        \midrule
        \multirow{6}{*}{\shortstack[l]{Equal\\compute}}
        & & & & 6.7 GPU-h & 9.0 GPU-h & 13.5 GPU-h & 17.6 GPU-h \\
        \cmidrule(lr){5-8}
        & 2 & 384 & 12 & 61.88 & 61.59 & 61.03 & 60.62 \\
        & 4 & 384 & 12 & 61.83 & \textbf{61.50} & \textbf{60.94} & \textbf{60.43} \\
        & 2 & 192 & 6  & \textbf{61.81} & 61.60 & 61.17 & 60.83 \\
        & 4 & 192 & 6  & 62.08 & 61.74 & 61.13 & 60.67 \\
        \cmidrule(lr){2-8}
        & \multicolumn{3}{c}{No rater} & 63.25 & 62.90 & 62.23 & 61.69 \\
        \bottomrule
    \end{tabular}}
\end{table}

\begin{table}[t]
  \centering
  \caption{Compute-matched comparison of the most efficient rater 
  composed with timestep-level designs, for DiT-S/2 on ImageNet
  $256{\times}256$ resuming from the 400k checkpoint.}
  \label{tab:tiny_compute_matched}
  \resizebox{0.9\linewidth}{!}{
  \begin{tabular}{l l cccc}
      \toprule
      & & \multicolumn{4}{c}{ImageNet $256{\times}256$, FID-50K $\downarrow$} \\
      \cmidrule(lr){3-6}
      & Method & 6.7 GPU-h & 9.0 GPU-h & 13.5 GPU-h & 17.6 GPU-h \\
      \midrule
      Baseline
      & Vanilla        & 63.25 & 62.90 & 62.23 & 61.69 \\
      \midrule
      \multirow{3}{*}{\shortstack[l]{Timestep-level}}
      & Min-SNR        & 62.15 & 61.35 & 60.11 & 59.05 \\
      & P2             & 61.35 & 60.25 & 58.77 & 57.45 \\
      & Logit-normal   & 60.68 & 59.64 & 58.06 & 57.16 \\
      \midrule
      \multirow{4}{*}{\shortstack[l]{+ NoiseRater\\($p=4$,$h=384$,$d=12$)}}
      & Vanilla        & 61.83 & 61.50 & 60.94 & 60.43 \\
      & Min-SNR        & 60.86 & 60.11 & 58.98 & 58.19 \\
      & P2             & 60.06 & 59.13 & 57.68 & 56.61 \\
      & Logit-normal   & \textbf{59.87} & \textbf{58.80} & \textbf{57.39} & \textbf{56.46} \\
      \bottomrule
  \end{tabular}}
\end{table}

\paragraph{Effect of rater size.}
Table~\ref{tab:effect_of_rater_size_compute} varies the rater's patch size $p$, hidden dimension $h$, and depth $d$. At equal steps, valuation quality grows with rater capacity: the largest rater ($p{=}2,h{=}384,d{=}12$) is best at every checkpoint, and halving $h$ and $d$ costs about $0.65$ FID at +80k, whereas doubling $p$ costs only $0.24$. At equal compute, the picture changes because a larger $p$ cuts the token count per stream by four and makes scoring much cheaper: the $p{=}4,h{=}384,d{=}12$ variant gives the best FID per GPU-hour, while the two shallow raters fall behind despite their lower cost. The most efficient rater is therefore not the smallest one but a full-capacity rater with coarser tokenization; we use it for the compute-matched comparison below.

\paragraph{Compute-matched comparison with timestep-level methods.}
Table~\ref{tab:tiny_compute_matched} composes the $p=4$ rater with the three timestep-level schemes of Section~\ref{sec:experiment_complement} under equal compute. Adding \emph{NoiseRater} improves every scheme at every budget. The reductions at 17.6 GPU-h ($-1.26$ on vanilla, $-0.86$ Min-SNR, $-0.84$ P2, $-0.70$ logit-normal) are smaller than the equal-step gains in Table~\ref{tab:timestep_composition}, as expected once scoring overhead is charged, but consistent in sign and comparable in relative size across schemes. {Appx.~\ref{app:seeds} reports this comparison over independent runs.}

\section{Conclusion}
We asked whether all noise realizations are equally useful for training diffusion models, and introduced \emph{NoiseRater}, a meta-learned rater that makes the utility of an individual noise instance measurable. Using it as a probe, we found that noise realizations at the same timestep are not equally useful, that their utility is a property of the noise in context rather than of the noise alone, and that selecting noise by utility is complementary to timestep-level weighting and scheduling. As a practical consequence, a frozen rater improves FID at matched compute and transfers across backbones. We hope these findings establish the choice of noise realization, alongside the choice of timestep, as an axis of diffusion training worth designing for.

\section*{AI Use Statement}
In this work, we used generative AI tools for the following tasks with required disclosure: implementing the Mixflow-MG~\citep{kemaev2025scalable} for keeping the memory cost of the unrolled computation tractable. We have not used generative AI tools to generate synthetic datasets, develop the theoretical framework, formulate mathematical claims, provide ingredients for or write the proofs, design the experimental methodology, interpret results, clean or reformat data; translation and qualitative or thematic data analysis are not applicable to this work. Additionally, we used generative AI tools for the following tasks with recommended disclosure: editing the paper to improve readability, suggesting the structure of the abstract and introduction, drafting parts of the results and conclusion sections from our tables and notes, and checking the consistency between the text and the reported numbers across tables. We have reviewed all AI-assisted work. All experiments, numbers, and tables were produced by the authors without AI assistance; every interpretive claim that originated from an LLM was checked by the authors against the reported results before inclusion; and all AI-edited text was read and approved by the authors. We take responsibility for the final content of this work, including text, claims or artifacts produced with the aid of generative AI.

\section*{Reproducibility Statement}
Code for the noise rater, the bilevel meta-training, and the rater-guided diffusion training is available at the anonymous repository linked in the abstract. The rater architecture and initialization are specified in Appx.~\ref{app:rater_arch_details}, all meta-training hyperparameters in Appx.~\ref{app:hyperparameters} (Table~\ref{tab:rater_hyperparams}), and compute requirements in Appx.~\ref{app:compute_resource}. The training and evaluation protocol, including checkpoints, step counts, and the FID-50K setup, is described in Section~\ref{sec:experiment_setup}; the checkpoint-matching rule for larger backbones is given in Appx.~\ref{app:larger_backbone_generalization}.


\bibliography{cite}

@article{he2025scaling,
  title={Scaling image and video generation via test-time evolutionary search},
  author={He, Haoran and Liang, Jiajun and Wang, Xintao and Wan, Pengfei and Zhang, Di and Gai, Kun and Pan, Ling},
  journal={arXiv preprint arXiv:2505.17618},
  year={2025}
}

@article{karras2022elucidating,
  title={Elucidating the design space of diffusion-based generative models},
  author={Karras, Tero and Aittala, Miika and Aila, Timo and Laine, Samuli},
  journal={Advances in neural information processing systems},
  volume={35},
  pages={26565--26577},
  year={2022}
}

@inproceedings{bechtle2021meta,
  title={Meta learning via learned loss},
  author={Bechtle, Sarah and Molchanov, Artem and Chebotar, Yevgen and Grefenstette, Edward and Righetti, Ludovic and Sukhatme, Gaurav and Meier, Franziska},
  booktitle={2020 25th International Conference on Pattern Recognition (ICPR)},
  pages={4161--4168},
  year={2021},
  organization={IEEE}
}

@article{sun2025noise,
  title={Is noise conditioning necessary for denoising generative models?},
  author={Sun, Qiao and Jiang, Zhicheng and Zhao, Hanhong and He, Kaiming},
  journal={arXiv preprint arXiv:2502.13129},
  year={2025}
}

@inproceedings{jiang2018mentornet,
  title={Mentornet: Learning data-driven curriculum for very deep neural networks on corrupted labels},
  author={Jiang, Lu and Zhou, Zhengyuan and Leung, Thomas and Li, Li-Jia and Fei-Fei, Li},
  booktitle={International conference on machine learning},
  pages={2304--2313},
  year={2018},
  organization={PMLR}
}

@inproceedings{wang2020training,
  title={Training noise-robust deep neural networks via meta-learning},
  author={Wang, Zhen and Hu, Guosheng and Hu, Qinghua},
  booktitle={Proceedings of the IEEE/CVF conference on computer vision and pattern recognition},
  pages={4524--4533},
  year={2020}
}

@article{ahn2024noise,
  title={A noise is worth diffusion guidance},
  author={Ahn, Donghoon and Kang, Jiwon and Lee, Sanghyun and Min, Jaewon and Kim, Minjae and Jang, Wooseok and Cho, Hyoungwon and Paul, Sayak and Kim, SeonHwa and Cha, Eunju and others},
  journal={arXiv preprint arXiv:2412.03895},
  year={2024}
}

@article{kim2025inference,
  title={Inference-time scaling for flow models via stochastic generation and rollover budget forcing},
  author={Kim, Jaihoon and Yoon, Taehoon and Hwang, Jisung and Sung, Minhyuk},
  journal={arXiv preprint arXiv:2503.19385},
  year={2025}
}

@article{ren2025driftlite,
  title={Driftlite: Lightweight drift control for inference-time scaling of diffusion models},
  author={Ren, Yinuo and Gao, Wenhao and Ying, Lexing and Rotskoff, Grant M and Han, Jiequn},
  journal={arXiv preprint arXiv:2509.21655},
  year={2025}
}

@inproceedings{maclaurin2015gradient,
  title={Gradient-based hyperparameter optimization through reversible learning},
  author={Maclaurin, Dougal and Duvenaud, David and Adams, Ryan},
  booktitle={International conference on machine learning},
  pages={2113--2122},
  year={2015},
  organization={PMLR}
}

@article{engstrom2025optimizing,
  title={Optimizing ml training with metagradient descent},
  author={Engstrom, Logan and Ilyas, Andrew and Chen, Benjamin and Feldmann, Axel and Moses, William and Madry, Aleksander},
  journal={arXiv preprint arXiv:2503.13751},
  year={2025}
}

@article{watson2023novo,
  title={De novo design of protein structure and function with RFdiffusion},
  author={Watson, Joseph L and Juergens, David and Bennett, Nathaniel R and Trippe, Brian L and Yim, Jason and Eisenach, Helen E and Ahern, Woody and Borst, Andrew J and Ragotte, Robert J and Milles, Lukas F and others},
  journal={Nature},
  volume={620},
  number={7976},
  pages={1089--1100},
  year={2023},
  publisher={Nature Publishing Group UK London}
}

@inproceedings{rombach2022high,
  title={High-resolution image synthesis with latent diffusion models},
  author={Rombach, Robin and Blattmann, Andreas and Lorenz, Dominik and Esser, Patrick and Ommer, Bj{\"o}rn},
  booktitle={Proceedings of the IEEE/CVF conference on computer vision and pattern recognition},
  pages={10684--10695},
  year={2022}
}

@article{polyak2024movie,
  title={Movie gen: A cast of media foundation models},
  author={Polyak, Adam and Zohar, Amit and Brown, Andrew and Tjandra, Andros and Sinha, Animesh and Lee, Ann and Vyas, Apoorv and Shi, Bowen and Ma, Chih-Yao and Chuang, Ching-Yao and others},
  journal={arXiv preprint arXiv:2410.13720},
  year={2024}
}

@article{ho2020denoising,
  title={Denoising diffusion probabilistic models},
  author={Ho, Jonathan and Jain, Ajay and Abbeel, Pieter},
  journal={Advances in neural information processing systems},
  volume={33},
  pages={6840--6851},
  year={2020}
}

@inproceedings{sohl2015deep,
  title={Deep unsupervised learning using nonequilibrium thermodynamics},
  author={Sohl-Dickstein, Jascha and Weiss, Eric and Maheswaranathan, Niru and Ganguli, Surya},
  booktitle={International conference on machine learning},
  pages={2256--2265},
  year={2015},
  organization={pmlr}
}

@article{ho2022classifier,
  title={Classifier-free diffusion guidance},
  author={Ho, Jonathan and Salimans, Tim},
  journal={arXiv preprint arXiv:2207.12598},
  year={2022}
}

@inproceedings{hang2025improved,
  title={Improved noise schedule for diffusion training},
  author={Hang, Tiankai and Gu, Shuyang and Bao, Jianmin and Wei, Fangyun and Chen, Dong and Geng, Xin and Guo, Baining},
  booktitle={Proceedings of the IEEE/CVF International Conference on Computer Vision},
  pages={4796--4806},
  year={2025}
}

@article{raya2026information,
  title={Information-Guided Noise Allocation for Efficient Diffusion Training},
  author={Raya, Gabriel and Nguyen, Bac and Batzolis, Georgios and Takida, Yuhta and Stancevic, Dejan and Murata, Naoki and Lai, Chieh-Hsin and Mitsufuji, Yuki and Ambrogioni, Luca},
  journal={arXiv preprint arXiv:2602.18647},
  year={2026}
}

@article{sun2026variance,
  title={Variance-Aware Adaptive Weighting for Diffusion Model Training},
  author={Sun, Nanlong and Shi, Lei},
  journal={arXiv preprint arXiv:2603.10391},
  year={2026}
}

@article{song2020score,
  title={Score-based generative modeling through stochastic differential equations},
  author={Song, Yang and Sohl-Dickstein, Jascha and Kingma, Diederik P and Kumar, Abhishek and Ermon, Stefano and Poole, Ben},
  journal={arXiv preprint arXiv:2011.13456},
  year={2020}
}

@article{li2025dynamic,
  title={Dynamic Search for Inference-Time Alignment in Diffusion Models},
  author={Li, Xiner and Uehara, Masatoshi and Su, Xingyu and Scalia, Gabriele and Biancalani, Tommaso and Regev, Aviv and Levine, Sergey and Ji, Shuiwang},
  journal={arXiv preprint arXiv:2503.02039},
  year={2025}
}

@article{singhal2025general,
  title={A general framework for inference-time scaling and steering of diffusion models},
  author={Singhal, Raghav and Horvitz, Zachary and Teehan, Ryan and Ren, Mengye and Yu, Zhou and McKeown, Kathleen and Ranganath, Rajesh},
  journal={arXiv preprint arXiv:2501.06848},
  year={2025}
}

@article{song2020denoising,
  title={Denoising diffusion implicit models},
  author={Song, Jiaming and Meng, Chenlin and Ermon, Stefano},
  journal={arXiv preprint arXiv:2010.02502},
  year={2020}
}

@article{calian2025datarater,
  title={Datarater: Meta-learned dataset curation},
  author={Calian, Dan A and Farquhar, Gregory and Kemaev, Iurii and Zintgraf, Luisa M and Hessel, Matteo and Shar, Jeremy and Oh, Junhyuk and Gy{\"o}rgy, Andr{\'a}s and Schaul, Tom and Dean, Jeffrey and others},
  journal={arXiv preprint arXiv:2505.17895},
  year={2025}
}

@inproceedings{tang2024tuning,
  title={Tuning-free alignment of diffusion models with direct noise optimization},
  author={Tang, Zhiwei and Peng, Jiangweizhi and Tang, Jiasheng and Hong, Mingyi and Wang, Fan and Chang, Tsung-Hui},
  booktitle={ICML 2024 Workshop on Structured Probabilistic Inference $\{$$\backslash$\&$\}$ Generative Modeling},
  year={2024}
}

@inproceedings{karunratanakul2024optimizing,
  title={Optimizing diffusion noise can serve as universal motion priors},
  author={Karunratanakul, Korrawe and Preechakul, Konpat and Aksan, Emre and Beeler, Thabo and Suwajanakorn, Supasorn and Tang, Siyu},
  booktitle={Proceedings of the IEEE/CVF Conference on Computer Vision and Pattern Recognition},
  pages={1334--1345},
  year={2024}
}

@inproceedings{guo2024initno,
  title={Initno: Boosting text-to-image diffusion models via initial noise optimization},
  author={Guo, Xiefan and Liu, Jinlin and Cui, Miaomiao and Li, Jiankai and Yang, Hongyu and Huang, Di},
  booktitle={Proceedings of the IEEE/CVF Conference on Computer Vision and Pattern Recognition},
  pages={9380--9389},
  year={2024}
}

@inproceedings{zhou2025golden,
  title={Golden noise for diffusion models: A learning framework},
  author={Zhou, Zikai and Shao, Shitong and Bai, Lichen and Zhang, Shufei and Xu, Zhiqiang and Han, Bo and Xie, Zeke},
  booktitle={Proceedings of the IEEE/CVF International Conference on Computer Vision},
  pages={17688--17697},
  year={2025}
}

@inproceedings{chen2024find,
  title={Find: Fine-tuning initial noise distribution with policy optimization for diffusion models},
  author={Chen, Changgu and Yang, Libing and Yang, Xiaoyan and Chen, Lianggangxu and He, Gaoqi and Wang, Changbo and Li, Yang},
  booktitle={Proceedings of the 32nd ACM International Conference on Multimedia},
  pages={6735--6744},
  year={2024}
}

@inproceedings{hang2023efficient,
  title={Efficient diffusion training via min-snr weighting strategy},
  author={Hang, Tiankai and Gu, Shuyang and Li, Chen and Bao, Jianmin and Chen, Dong and Hu, Han and Geng, Xin and Guo, Baining},
  booktitle={Proceedings of the IEEE/CVF international conference on computer vision},
  pages={7441--7451},
  year={2023}
}

@article{kim2024denoising,
  title={Denoising task difficulty-based curriculum for training diffusion models},
  author={Kim, Jin-Young and Go, Hyojun and Kwon, Soonwoo and Kim, Hyun-Gyoon},
  journal={arXiv preprint arXiv:2403.10348},
  year={2024}
}

@inproceedings{peebles2023scalable,
  title={Scalable diffusion models with transformers},
  author={Peebles, William and Xie, Saining},
  booktitle={Proceedings of the IEEE/CVF international conference on computer vision},
  pages={4195--4205},
  year={2023}
}

@inproceedings{choi2022perception,
  title={Perception prioritized training of diffusion models},
  author={Choi, Jooyoung and Lee, Jungbeom and Shin, Chaehun and Kim, Sungwon and Kim, Hyunwoo and Yoon, Sungroh},
  booktitle={Proceedings of the IEEE/CVF conference on computer vision and pattern recognition},
  pages={11472--11481},
  year={2022}
}

@inproceedings{ren2018learning,
  title={Learning to reweight examples for robust deep learning},
  author={Ren, Mengye and Zeng, Wenyuan and Yang, Bin and Urtasun, Raquel},
  booktitle={International conference on machine learning},
  pages={4334--4343},
  year={2018},
  organization={PMLR}
}

@article{shu2019meta,
  title={Meta-weight-net: Learning an explicit mapping for sample weighting},
  author={Shu, Jun and Xie, Qi and Yi, Lixuan and Zhao, Qian and Zhou, Sanping and Xu, Zongben and Meng, Deyu},
  journal={Advances in neural information processing systems},
  volume={32},
  year={2019}
}

@inproceedings{esser2024scaling,
  title={Scaling rectified flow transformers for high-resolution image synthesis},
  author={Esser, Patrick and Kulal, Sumith and Blattmann, Andreas and Entezari, Rahim and M{\"u}ller, Jonas and Saini, Harry and Levi, Yam and Lorenz, Dominik and Sauer, Axel and Boesel, Frederic and others},
  booktitle={Forty-first international conference on machine learning},
  year={2024}
}

@article{wang2024evaluating,
  title={Evaluating the design space of diffusion-based generative models},
  author={Wang, Yuqing and He, Ye and Tao, Molei},
  journal={Advances in Neural Information Processing Systems},
  volume={37},
  pages={19307--19352},
  year={2024}
}

@inproceedings{karras2019style,
  title={A style-based generator architecture for generative adversarial networks},
  author={Karras, Tero and Laine, Samuli and Aila, Timo},
  booktitle={Proceedings of the IEEE/CVF conference on computer vision and pattern recognition},
  pages={4401--4410},
  year={2019}
}

@inproceedings{deng2009imagenet,
  title={Imagenet: A large-scale hierarchical image database},
  author={Deng, Jia and Dong, Wei and Socher, Richard and Li, Li-Jia and Li, Kai and Fei-Fei, Li},
  booktitle={2009 IEEE conference on computer vision and pattern recognition},
  pages={248--255},
  year={2009},
  organization={Ieee}
}

@article{heusel2017gans,
  title={Gans trained by a two time-scale update rule converge to a local nash equilibrium},
  author={Heusel, Martin and Ramsauer, Hubert and Unterthiner, Thomas and Nessler, Bernhard and Hochreiter, Sepp},
  journal={Advances in neural information processing systems},
  volume={30},
  year={2017}
}

@article{kemaev2025scalable,
  title={Scalable meta-learning via mixed-mode differentiation},
  author={Kemaev, Iurii and Calian, Dan A and Zintgraf, Luisa M and Farquhar, Gregory and Van Hasselt, Hado},
  journal={arXiv preprint arXiv:2505.00793},
  year={2025}
}

@inproceedings{recht2019imagenet,
  title={Do imagenet classifiers generalize to imagenet?},
  author={Recht, Benjamin and Roelofs, Rebecca and Schmidt, Ludwig and Shankar, Vaishaal},
  booktitle={International conference on machine learning},
  pages={5389--5400},
  year={2019},
  organization={PMLR}
}
\bibliographystyle{abbrvnat}

\appendix
\newpage
\section{Related Work}

\paragraph{Noise Optimization at Inference Time.}
Recent work demonstrates that optimizing initial noise at inference time can significantly improve generation quality. One line of research directly optimizes the noise latent through backpropagation across denoising steps. These methods iteratively update the noise to maximize human-preference rewards~\citep{tang2024tuning}, satisfy specific motion constraints~\citep{karunratanakul2024optimizing}, or align with internal attention scores~\citep{guo2024initno}. While effective and training-free, they incur massive computational overhead at inference due to repeated denoising loops. To amortize this cost, other approaches train auxiliary models to directly predict optimal noise. For instance, FIND~\citep{chen2024find} uses reinforcement learning to adjust the mean and variance of the initial Gaussian, while GoldenNoise~\citep{zhou2025golden} trains a network to predict high-quality noise perturbations. Crucially, all treat noise as a test-time control variable. In contrast, we explore the untapped potential of evaluating and weighting noise during training.

\paragraph{Training-Time Loss Reweighting and Noise Selection.}
Various loss reweighting and noise scheduling strategies improve diffusion training efficiency. Most methods operate at the \emph{timestep level}, adjusting the relative importance or sampling frequency of different noise magnitudes. For loss reweighting, P2~\citep{choi2022perception} prioritizes highly corrupted timesteps to learn global concepts, Min-SNR~\citep{hang2023efficient} balances inter-timestep optimization conflicts, and \citet{sun2026variance} dynamically adjusts weights based on the variance of loss distributions across log-SNR levels. Concurrently, noise scheduling approaches optimize the distribution of sampled timesteps: \citet{hang2025improved} introduces a fixed schedule focusing on the $\log \text{SNR} \approx 0$ region. \citet{kim2024denoising} uses curriculum learning to progressively introduce harder timesteps, and \citet{raya2026information} proposes a dynamic, information-guided allocation based on entropy reduction.~\citet{wang2024evaluating} theoretically analyzes different total weighting, time, and noise schedules. In addition, EDM~\citep{karras2022elucidating} provides a unifying perspective on diffusion design, showing that choices of noise parameterization, preconditioning, and loss scaling implicitly define effective weighting across noise levels. However, these methods treat all noise instances at a given timestep as equally informative.
Recent work has also begun to question more fundamental assumptions about noise in diffusion models. For example,~\citet{sun2025noise} shows that explicit noise conditioning may not be strictly necessary, with models often maintaining competitive performance even without access to timestep information. While such work investigates whether noise information is required at all, it does not address how the variability of noise \emph{instances} influences learning. In contrast, we assume standard conditioning and focus on a complementary question: \emph{given noise, which specific realizations are most useful for training?} 

\paragraph{Meta-Learning for Training Optimization.} Meta-learning techniques have been explored for adaptive data weighting~\citep{bechtle2021meta}. Early methods~\citep{ren2018learning,jiang2018mentornet} used online meta-gradients to dynamically weight training examples. To improve scalability and generalization, subsequent work parameterized these weighting mechanisms using neural networks trained via bilevel optimization, evolving from dynamic loss-to-weight mappings~\citep{shu2019meta,wang2020training} to large-scale dataset curation models~\citep{calian2025datarater}. We introduce this bilevel optimization paradigm to diffusion models for the first time. Crucially, we shift the fundamental object of valuation: instead of rating the quality of training \emph{data}, our meta-network is designed to rate the learning utility of the injected \emph{noise} instances.

\section{Mathematical Analysis}

\subsection{Proof of Thm.~\ref{thm:meta_gradient}}
Define $F(\theta,\eta) := \nabla_\theta \mathcal{L}_{\mathrm{inner}}(\theta;\eta)$.
At the inner optimum $\theta^*(\eta)$, the first-order optimality condition gives $F(\theta^*(\eta),\eta) = \nabla_\theta \mathcal{L}_{\mathrm{inner}}(\theta^*(\eta);\eta) = 0$.
Since $\mathcal{L}_{\mathrm{inner}}(\theta;\eta)$ is $\mu$-strongly convex in $\theta$, its Hessian $H(\eta) := \nabla_\theta^2 \mathcal{L}_{\mathrm{inner}}(\theta^*(\eta);\eta)$ is positive definite and hence invertible. Therefore, by the implicit function theorem, $\theta^*(\eta)$ is locally unique and differentiable with respect to $\eta$.
Differentiating the optimality condition $\nabla_\theta \mathcal{L}_{\mathrm{inner}}(\theta^*(\eta);\eta)=0$ with respect to $\eta$ yields $\nabla_\theta^2 \mathcal{L}_{\mathrm{inner}}(\theta^*(\eta);\eta) \frac{\partial \theta^*(\eta)}{\partial \eta} + \nabla_\eta \nabla_\theta \mathcal{L}_{\mathrm{inner}}(\theta^*(\eta);\eta) = 0$.
Solving for $\partial \theta^*(\eta)/\partial \eta$, we obtain $\frac{\partial \theta^*(\eta)}{\partial \eta} = - \Bigl(\nabla_\theta^2 \mathcal{L}_{\mathrm{inner}}(\theta^*(\eta);\eta)\Bigr)^{-1} \nabla_\eta \nabla_\theta \mathcal{L}_{\mathrm{inner}}(\theta^*(\eta);\eta)$.
Now define the outer objective $J(\eta) = \mathcal{L}_{\mathrm{val}}(\theta^*(\eta))$.
By the chain rule, $\nabla_\eta J(\eta) = \left(\frac{\partial \theta^*(\eta)}{\partial \eta}\right)^\top \nabla_\theta \mathcal{L}_{\mathrm{val}}(\theta^*(\eta))$.
Substituting the expression above gives $\nabla_\eta J(\eta) = - \Bigl(\nabla_\eta \nabla_\theta \mathcal{L}_{\mathrm{inner}}(\theta^*(\eta);\eta)\Bigr)^\top \Bigl(\nabla_\theta^2 \mathcal{L}_{\mathrm{inner}}(\theta^*(\eta);\eta)\Bigr)^{-1} \nabla_\theta \mathcal{L}_{\mathrm{val}}(\theta^*(\eta))$.
This proves the claimed meta-gradient formula.

\subsection{Proof of Thm.~\ref{thm:alignment}}
Let $g_k(\theta) := \nabla_\theta \ell^{(k)}(\theta)$ and $g_{\mathrm{val}}(\theta) := \nabla_\theta \mathcal{L}_{\mathrm{val}}(\theta)$.
The single inner update is $\theta^+(\eta) = \theta - \alpha  \sum_{k=1}^K w^{(k)}(\eta) g_k(\theta)$, hence $\theta^+(\eta)-\theta = -\alpha  \sum_{k=1}^K w^{(k)}(\eta) g_k(\theta)$.
Since $\mathcal{L}_{\mathrm{val}}$ has $L$-Lipschitz continuous gradients, the descent lemma gives $\mathcal{L}_{\mathrm{val}}(\theta^+) \leq \mathcal{L}_{\mathrm{val}}(\theta) + \left\langle g_{\mathrm{val}}(\theta), \theta^+ - \theta \right\rangle + \frac{L}{2}\|\theta^+ - \theta\|^2$.
Substituting the update direction, the linear term becomes $\left\langle g_{\mathrm{val}}(\theta), \theta^+ - \theta \right\rangle = -\alpha  \sum_{k=1}^K w^{(k)}(\eta) \left\langle g_{\mathrm{val}}(\theta), g_k(\theta) \right\rangle$, and $\frac{L}{2}\|\theta^+ - \theta\|^2 = \frac{L\alpha^2}{2} \left\|\sum_{k=1}^K w^{(k)}(\eta) g_k(\theta)\right\|^2 = \mathcal{O}(\alpha^2)$.
Therefore, $\mathcal{L}_{\mathrm{val}}(\theta^+(\eta)) \leq \mathcal{L}_{\mathrm{val}}(\theta) - \alpha  \sum_{k=1}^K w^{(k)}(\eta) \left\langle \nabla_\theta \mathcal{L}_{\mathrm{val}}(\theta), \nabla_\theta \ell^{(k)}(\theta) \right\rangle + \mathcal{O}(\alpha^2)$.
Thus, up to second-order terms in $\alpha$, decreasing the validation loss favors larger weights on noise instances whose training gradients have larger positive alignment with the validation gradient.

\section{Experimental Details}

\subsection{Rater Architecture}
\label{app:rater_arch_details}
\begin{figure}[ht]
    \centering
    \includegraphics[width=1.0\linewidth]{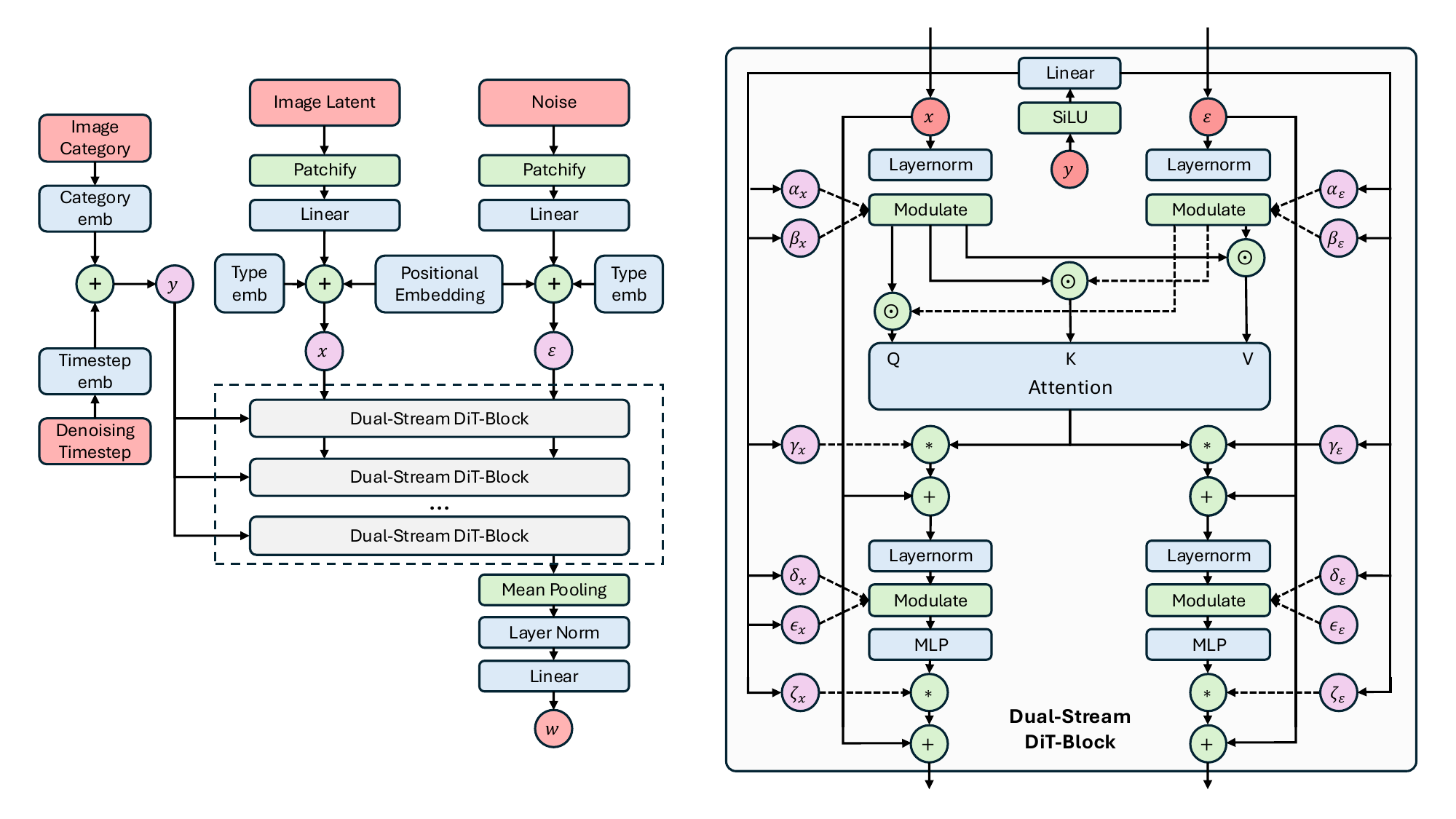}
    \caption{The illustration of the noise rater's architecture, which contains multiple dual-stream DiT blocks to incorporate several condition inputs.}
    \label{fig:rater_arch}
\end{figure}

\paragraph{Inputs and tokenization.} The rater takes as input a clean image latent $x_0$, a noise sample $\epsilon$, a diffusion timestep $t$, and a class label $c$. We operate in the latent space of the publicly available \texttt{stabilityai/sd-vae-ft-mse} VAE~\citep{rombach2022high}, which encodes a $256{\times}256$ RGB image into a $4 \times 32 \times 32$ latent ($8\times$ spatial downsampling, $4$ channels); the noise sample $\epsilon$ is drawn from $\mathcal{N}(0, I)$ with the same shape. Both $x_0$ and $\epsilon$ are patchified by two independent $\textsc{PatchEmbed}$ projections (Conv2d with kernel size = stride = $p = 2$), producing two token sequences of length $T = (32/p)^2 = 256$, each in $\mathbb{R}^{T \times D}$ where $D = 384$ is the hidden size. Both streams share a single frozen 2D sin-cos positional embedding (added before the first block) and additionally receive learnable per-stream type embeddings that mark their stream identity, so the joint attention can distinguish image tokens from noise tokens.

\paragraph{Conditioning.} The conditioning vector $y \in \mathbb{R}^{D}$ is the sum of (i) a sinusoidal timestep embedding passed through a two-layer MLP and (ii) a learnable class embedding. The same $y$ is broadcast to every block.

\paragraph{Dual-stream DiT block.} Each block maintains two independent residual paths, one for the image stream and one for the noise stream. Each path has their own LayerNorm, MLP, and adaLN modulation parameters. Following DiT, each stream produces six modulation values per block; concretely, a single Linear layer projects $y$ to $12D$ outputs and chunks them into the twelve required scalars per token dimension. Attention is the only sublayer shared across streams: at each block we form the concatenated sequence $[\hat{x}_0^{(\ell)}; \hat{\epsilon}^{(\ell)}] \in \mathbb{R}^{2T \times D}$ from the modulated normalized tokens of the two streams, run a single multi-head self-attention, and split the output back into per-stream halves before applying the per-stream gates and residuals. Joint attention lets noise tokens directly query image content (and vice versa) without requiring additional cross-attention parameters.

\paragraph{Output head.} After the final block, the image-stream tokens are discarded; the noise-stream tokens are mean-pooled across the spatial dimension, layer-normalized, and projected by a single linear layer to a scalar score $w \in \mathbb{R}$. 

\paragraph{Hyperparameters and initialization.} We use the same per-block configuration as DiT-S/2: depth $= 12$, hidden size $D = 384$, $6$ attention heads, MLP ratio $= 4$, patch size $p = 2$, operating on $32 \times 32$ latents (i.e., $T = 256$ tokens per stream). The total parameter count is approximately $\sim$$60$M, roughly twice that of DiT-S/2 because of the duplicated MLPs and modulation networks.

Unlike generative DiT models, we do \emph{not} adopt adaLN-Zero (i.e., zero-initializing the adaLN projection and the output head); instead all linear layers use Xavier initialization. We found this necessary because the rater is trained via bilevel optimization: each rater update requires running an inner loop of diffusion updates whose loss is reweighted by the rater's scores. Under adaLN-Zero, the rater's score head is zero-initialized, so all candidate noises receive identical (zero) scores at initialization. After group-wise normalization, this collapses the inner-loop loss weights to a constant, yielding zero gradient signal back to the rater and stalling training before it begins. Standard Xavier initialization breaks this symmetry by giving the rater a non-trivial starting function, and yields stable training across all settings reported in the main paper.

\subsection{Training Hyperparameters} 
\label{app:hyperparameters}
Tab.~\ref{tab:rater_hyperparams} lists the hyperparameters used for noise rater training. The rater is trained on top of a frozen DiT-S/2 checkpoint via a meta-learning objective: an inner loop takes $5$ gradient steps on the diffusion model with rater-weighted noise, and an outer loop updates the rater to minimize the resulting diffusion loss on a held-out validation split.

\begin{table}[t]
    \centering
    \caption{Hyperparameters for noise rater training.}
    \begin{tabular}{ll}
        \toprule
        Hyperparameter & Value \\
        \midrule
        Candidate pool size                   & 4 \\
        Outer (rater) learning rate           & $1 \times 10^{-4}$ \\
        Inner (diffusion) learning rate       & $5 \times 10^{-2}$ \\
        Inner steps per meta update           & 5 \\
        Meta refresh steps                    & 4 \\
        Total meta-training steps             & 2{,}000 \\
        Inner model batch size                & 32 \\
        Validation batch size                 & 128 \\
        Gradient clip norm                    & 1.0 \\
        \bottomrule
    \end{tabular}
    \label{tab:rater_hyperparams}
\end{table}

\subsection{Computation Resources} \label{app:compute_resource}

\paragraph{Rater training.} With the hyperparameters in Tab.~\ref{tab:rater_hyperparams}, training the noise rater takes approximately $4$ hours of wall-clock time on $4$ NVIDIA H100 GPUs (16 CPU cores). This is a one-time cost that is amortized across all subsequent diffusion runs.

\paragraph{Diffusion training.} All diffusion training is conducted on $4$ NVIDIA H100 GPUs.

\textit{(i) DiT-S/2 main runs.} Pre-training DiT-S/2 from scratch to $400$k steps takes about $22$ hours. Each $80$k-step continuation run takes $4.3$, $5.6$, $7.0$, and $9.0$ hours for $k{=}1, 2, 4, 8$ respectively.

\textit{(ii) Larger-backbone generalization.} (Appx.~\ref{app:larger_backbone_generalization}) On DiT-B/2, the $80$k continuation takes $5.4$ hours without the rater and $6.7$ hours with the rater at $k{=}2$. On DiT-L/2, it takes $9.2$ hours without the rater and $10.4$ hours with the rater at $k{=}2$.

\section{Rater of different training stage.}
We explored whether the rater is equally beneficial when applied at different stages of diffusion training. We test this by training a rater on top of DiT-S/2 checkpoints at 100k, 400k, and 500k steps, and then continuing diffusion training for an additional 80k steps with $k{=}4$. As shown in Tab.~\ref{tab:training_stage}, the rater consistently improves FID at every stage and every checkpoint, but the magnitude of the gain varies substantially: applied at 400k, the rater yields a $2.7\%$--$3.0\%$ relative FID reduction, whereas at 100k and 500k the gain shrinks to $0.5\%$--$1.3\%$ and $0.4\%$--$0.7\%$ respectively. We attribute this gap to hyperparameter sensitivity. All three runs reuse the same training recipe tuned on the 400k checkpoint; we did not re-tune for the early or late training regimes, where the diffusion model's noise-to-signal characteristics differ noticeably from those at 400k. We expect stage-specific tuning to recover most of the gap, and we leave a thorough hyperparameter sweep across training stages to future work. Even without such tuning, the rater never hurts and delivers consistent improvements across the full training trajectory.

\begin{table}[h!]
    \centering
    \small
    \caption{Effect of applying the rater at different stages of DiT-S/2 training on ImageNet $256{\times}256$. We train a rater on top of the 100k, 400k, and 500k checkpoints, respectively, then continue diffusion training for an additional 80k steps. FID-50K is reported every 20k steps.}
    \label{tab:training_stage}
    \begin{tabular}{l cc cc cc}
        \toprule
        & \multicolumn{2}{c}{100k} & \multicolumn{2}{c}{400k} & \multicolumn{2}{c}{500k} \\
        \cmidrule(lr){2-3} \cmidrule(lr){4-5} \cmidrule(lr){6-7}
        Train steps & w/o rater & w/i rater & w/o rater & w/i rater & w/o rater & w/i rater \\
        \midrule
        +20k & 86.08 & \textbf{85.09} & 63.60 & \textbf{61.88} & 60.42 & \textbf{60.13} \\
        +40k & 82.39 & \textbf{81.33} & 62.93 & \textbf{61.03} & 59.96 & \textbf{59.54} \\
        +60k & 79.35 & \textbf{78.57} & 62.26 & \textbf{60.37} & 59.38 & \textbf{58.98} \\
        +80k & 76.63 & \textbf{76.28} & 61.69 & \textbf{59.97} & 58.90 & \textbf{58.68} \\
        \bottomrule
    \end{tabular}
    \vspace{-1em}
\end{table}

\section{Generalization to larger backbones.} \label{app:larger_backbone_generalization}
{We test whether a rater trained on one backbone transfers to larger ones. The rater is trained on DiT-S/2 and applied, without retraining, to DiT-B/2 and DiT-L/2. It transfers to DiT-B/2, while its gain becomes negligible on DiT-L/2.}

\paragraph{Training stage matching.}

Since different backbones converge at different rates, naively reusing the same step count would compare backbones at very different points in their training trajectories. To match the rater at a suitable training stage, we instead select target checkpoints for DiT-B/2 and DiT-L/2 that match the \emph{absolute training loss} of DiT-S/2 at $400$k steps.

Concretely, let $t$ denote the training step, and let $L_S(t)$, $L_B(t)$, and $L_L(t)$ denote the training loss curves of DiT-S/2, DiT-B/2, and DiT-L/2 respectively. We define the matched checkpoint for each larger backbone as
\begin{equation}
    t^\star_B = \min \big\{\, t \,:\, \widetilde{L}_B(t') \le \widetilde{L}_S(400\text{k}) \;\;\text{for all}\;\; t' \in [t, t + W] \,\big\},
\end{equation}
and analogously for $t^\star_L$, where $\widetilde{L}(\cdot)$ denotes the training loss smoothed with a centered rolling window of size $500$, and $W = 1000$ logged steps is a stability window that prevents single-step dips from triggering a spurious match. The smoothing and stability window together make the selection robust to the noisy loss curves typical of diffusion training.

Applying this procedure with $L_S(400\text{k})$ as the target yields $t^\star_B = 211$k for DiT-B/2 and $t^\star_L = 119$k for DiT-L/2. Intuitively, larger backbones reach the same loss level in fewer steps, which is consistent with prior observations on capacity-driven sample efficiency in diffusion training~\citep{peebles2023scalable}. Thus, we take the rater trained on DiT-S/2 at the 400k checkpoint and apply it to train DiT-B/2 and DiT-L/2 from their 200k and 100k checkpoints (rounded it down to the nearest saved checkpoint), respectively, for an additional 80k steps with $k{=}2$.


\paragraph{Generalization performance.}
{As shown in Tab.~\ref{tab:larger_model_generalization}, the DiT-S/2 rater improves FID on DiT-B/2 at every checkpoint, by $0.83$ to $0.99$, with a gain of $0.85$ at +80k ($48.65 \rightarrow 47.80$). On DiT-L/2, the gain is $0.16$ to $0.27$ at the first three checkpoints and disappears at +80k ($36.95$ vs.\ $36.94$). A rater trained on a small backbone therefore transfers to a moderately larger one, but not to DiT-L/2.}

\begin{table}[t]
    \centering
    \caption{Generalization to larger backbones. A rater trained on DiT-S/2 (at the 400k checkpoint) is applied off-the-shelf to train DiT-B/2 and DiT-L/2 on ImageNet $256{\times}256$, resuming from their 200k and 100k checkpoints, respectively. The rater is frozen and never retrained on the larger backbones.}
    \label{tab:larger_model_generalization}
    \resizebox{0.7\linewidth}{!}{
    \begin{tabular}{lcccccc}
        \toprule
        & \multicolumn{2}{c}{} & \multicolumn{4}{c}{ImageNet $256{\times}256$, FID-50K $\downarrow$} \\
        \cmidrule(lr){4-7}
        Backbone & Resume from & Strategy & +20k & +40k & +60k & +80k \\
        \midrule
        \multirow{2}{*}{DiT-B/2} & \multirow{2}{*}{200k} & No Rater    & 53.54          & 51.78          & 50.15          & 48.65          \\
                                 &                       & Rater ($k{=}2$) & \textbf{52.55} & \textbf{50.93} & \textbf{49.32} & \textbf{47.80} \\
        \midrule
        \multirow{2}{*}{DiT-L/2} & \multirow{2}{*}{100k} & No Rater    & 48.09          & 43.60          & 40.13          & 36.95          \\
                                 &                       & Rater ($k{=}2$) & \textbf{47.93} & \textbf{43.34} & \textbf{39.86} & \textbf{36.94} \\
        \bottomrule
    \end{tabular}}
\vspace{-1em}
\end{table}

\section{Effect of the Number of Inner Steps $S$}\label{app:inner_steps}

The meta-objective $J_S$ in Section~\ref{sec:bilevel_meta_objective} is a finite-step truncation of the idealized bilevel objective $J$ analyzed in Thm.~\ref{thm:meta_gradient}. The number of inner steps $S$ therefore controls how closely the rater's training signal approximates the converged inner optimum, but also the cost of each meta-update, since the meta-gradient is obtained by differentiating through all $S$ unrolled diffusion updates. To quantify this trade-off, we train raters with $S \in \{1, 3, 5, 7\}$, keeping every other hyperparameter in Tab.~\ref{tab:rater_hyperparams} fixed, and evaluate each rater under the same Stage-II protocol as the main experiments (DiT-S/2, ImageNet $256\times256$, {class-conditional without CFG}, Top-1 selection with $K'=4$, $80$k continuation steps from the $400$k checkpoint).

\begin{table}[h]
\centering
\caption{Effect of the number of inner steps $S$ on meta-training cost and downstream quality. Meta-training cost is measured in GPU-hours on 4$\times$H100 for the full $2$k meta-training steps. FID-50K is reported at $+80$k Stage-II steps. $S=5$ is the setting used in all main-paper experiments.}
\label{tab:inner_steps}
\begin{tabular}{ccc}
\toprule
$S$ & Meta-training cost (GPU-h) & FID-50K $\downarrow$ \\
\midrule
1 & 3.87  & 60.01 \\
3 & 9.75  & 60.19 \\
5 & 15.82 & \textbf{59.97} \\
7 & 21.49 & 61.41 \\
\midrule
No rater & -- & 61.69 \\
\bottomrule
\end{tabular}
\end{table}

\paragraph{Cost.} Meta-training cost grows linearly in $S$, at roughly $3$ GPU-hours per additional inner step, because each meta-update must run $S$ diffusion forward/backward passes and then backpropagate through the resulting trajectory. Memory grows with $S$ as well: mixed-mode differentiation~\citep{kemaev2025scalable} avoids materializing second-order terms, but the activations of all $S$ inner steps must still be retained for the reverse pass.

\paragraph{Downstream quality.} Downstream FID is comparatively insensitive to $S$ in the range $1 \le S \le 5$: all three raters improve over vanilla training by $1.5$--$1.7$ FID, and $S=5$ is best by a small margin. Increasing to $S=7$ degrades FID to $61.41$, close to the no-rater baseline. We attribute this to the well-known instability of long unrolled meta-gradients~\citep{maclaurin2015gradient}: as the unroll length grows, the meta-gradient becomes higher-variance and more poorly conditioned, and with the meta-learning rate held fixed the rater fails to converge to a useful ranking within the $2$k-step budget. We did not re-tune the meta-learning rate for $S=7$.

\section{{Compute comparison including rater training}}
\label{app:meta_cost}

The equal-compute results of Section~\ref{sec:experiment_improve} count the per-step scoring cost but not the one-time rater training. Here we charge the rater training in full to a single continuation run and compare against vanilla training at the same total GPU-hours. All totals start from the shared 400k checkpoint, whose pre-training cost is common to both methods and excluded. We use the $p{=}4, h{=}384, d{=}12$ rater, the most efficient variant in Table~\ref{tab:effect_of_rater_size_compute}. With this rater, the continuation costs $4.85$ GPU-h per 20k steps, against $4.40$ for vanilla training.

\paragraph{Default setting ($S{=}5$).}
With $S{=}5$ inner steps, training the $p{=}4$ rater costs $11.6$ GPU-h. As shown in the upper block of Tab.~\ref{tab:meta_cost}, a single run that pays this cost in full trails vanilla training by $0.14$ to $0.25$ FID at the first three budgets, and matches it at about $31$ GPU-h ($60.21$ vs.\ $60.19$). With the default setting, noise selection therefore does not reduce the total compute of a single run.

\paragraph{A single inner step ($S{=}1$).}
Tab.~\ref{tab:inner_steps} shows that the downstream FID of the full-size rater is insensitive to $S$ between $1$ and $5$, while the cost of rater training grows linearly in $S$. We therefore train the $p{=}4$ rater with $S{=}1$. This reduces the cost of rater training from $11.6$ to $3.1$ GPU-h, and the FID at +80k is unchanged ($60.18$ for $S{=}1$ vs.\ $60.21$ for $S{=}5$). As shown in Tab.~\ref{tab:meta_cost}, with the rater training included, \emph{NoiseRater} with $S{=}1$ has lower FID than vanilla training at every budget. It reaches $60.18$ FID with $22.5$ GPU-h in total, while vanilla training needs $30.8$ GPU-h to reach $60.19$, i.e., \emph{NoiseRater} uses about $27\%$ less compute. The $S{=}1$ result comes from a single run; for reference, the standard deviation across runs for this rater is $0.13$ FID (Appx.~\ref{app:seeds}).

\begin{table}[h!]
  
  \centering
  \small
  \caption{{Compute comparison including rater training, for DiT-S/2 on ImageNet $256{\times}256$ resuming from the 400k checkpoint. The total compute of \emph{NoiseRater} includes the rater training ($11.6$ GPU-h for $S{=}5$, $3.1$ GPU-h for $S{=}1$) and the per-step scoring cost of the $p{=}4$ rater. Vanilla training is reported at the same total compute. $\Delta$ is the FID difference between \emph{NoiseRater} and vanilla training; negative values favor \emph{NoiseRater}.}}
  \label{tab:meta_cost}
  \begin{tabular}{l c c cc c}
      \toprule
      & & & \multicolumn{2}{c}{ImageNet $256{\times}256$, FID-50K $\downarrow$} & \\
      \cmidrule(lr){4-5}
      \textbf{Rater} & \textbf{Total GPU-h} & \textbf{Steps} & \textbf{NoiseRater} & \textbf{Vanilla} & $\Delta$ \\
      \midrule
      \multirow{4}{*}{\shortstack[l]{$S{=}5$\\(default)}}
          & 16.5 & +20k & 62.10 & \textbf{61.85} & $+0.25$ \\
          & 21.3 & +40k & 61.40 & \textbf{61.26} & $+0.14$ \\
          & 26.2 & +60k & 60.81 & \textbf{60.62} & $+0.19$ \\
          & 31.0 & +80k & 60.21 & \textbf{60.19} & $+0.02$ \\
      \midrule
      \multirow{4}{*}{$S{=}1$}
          & 7.9  & +20k & \textbf{61.89} & 63.05 & $-1.16$ \\
          & 12.8 & +40k & \textbf{61.34} & 62.33 & $-0.99$ \\
          & 17.6 & +60k & \textbf{60.73} & 61.70 & $-0.97$ \\
          & 22.5 & +80k & \textbf{60.18} & 61.11 & $-0.93$ \\
      \bottomrule
  \end{tabular}
\end{table}

\section[Repeatability across independent runs]{{Repeatability across independent runs}}
\label{app:seeds}

The results in the main paper come from a single run per configuration. We test whether the gains from noise selection exceed the run-to-run variation by repeating the diffusion training stage with independent seeds. All runs follow the protocol of Section~\ref{sec:experiment_setup} (DiT-S/2, ImageNet $256{\times}256$, class-conditional without CFG, top-1 selection with $k{=}4$, 80k continuation steps from the 400k checkpoint). We report the mean and the sample standard deviation of FID-50K.

\paragraph{Setup.}
All runs resume from the same 400k checkpoint, and the seed of the continuation run controls the order of the training data, the sampled timesteps, and the candidate noises. We repeat vanilla training and training with the full-size rater with three continuation seeds each. For the $p{=}4$ rater of Section~\ref{sec:experiment_improve}, we separate the two sources of randomness in our pipeline: we either keep the rater fixed and vary the continuation seed, or keep the continuation seed fixed and vary the seed of rater training. We also repeat the three timestep-level schemes of Section~\ref{sec:experiment_complement} with and without the $p{=}4$ rater.

\paragraph{Equal training steps.}
As shown in Tab.~\ref{tab:seeds_equal_steps}, the standard deviation across runs is between $0.02$ and $0.13$ FID for every configuration. The full-size rater reduces the mean FID of vanilla training by $1.71$, and the $p{=}4$ rater by $1.42$. Varying the seed of rater training gives nearly the same mean as varying the continuation seed ($60.24$ vs.\ $60.25$), so the gain is not tied to a particular rater. On top of the three timestep-level schemes, the $p{=}4$ rater reduces the mean FID by $1.21$ to $1.27$. In all cases, the gain is roughly an order of magnitude larger than the standard deviation across runs.

\paragraph{Equal compute.}
Tab.~\ref{tab:seeds_equal_compute} repeats the compute-matched comparison of Table~\ref{tab:tiny_compute_matched} at the 17.6 GPU-h budget. As in Section~\ref{sec:experiment_improve}, the budget counts the per-step scoring cost but not the one-time rater training. Adding \emph{NoiseRater} reduces the mean FID by $0.83$ to $1.19$, while the standard deviation across runs is at most $0.12$. The reductions are smaller than the equal-step ones in Tab.~\ref{tab:seeds_equal_steps}, as expected once scoring overhead is charged, but they remain several times larger than the run-to-run variation for every scheme.

\begin{table}[h!]
  
  \centering
  \small
  \caption{{Repeatability at equal training steps. FID-50K at +80k for DiT-S/2 on ImageNet $256{\times}256$, reported as mean $\pm$ standard deviation over three independent runs (two for Logit-normal + NoiseRater). $\Delta$ is the reduction in mean FID from adding NoiseRater.}}
  \label{tab:seeds_equal_steps}
  \begin{tabular}{l c c}
      \toprule
      & ImageNet $256{\times}256$, FID-50K $\downarrow$ & \\
      \cmidrule(lr){2-2}
      \textbf{Method} & \textbf{+80k} & $\Delta_{+80k}$ \\
      \midrule
      Vanilla                              & $61.67 \pm 0.02$ & -- \\
      \;\;+ NoiseRater (full)              & $\mathbf{59.96 \pm 0.03}$ & $-1.71$ \\
      \;\;+ NoiseRater ($p{=}4$),          & \multirow{2}{*}{$60.25 \pm 0.13$} & \multirow{2}{*}{$-1.42$} \\
      \;\;\;\;\textit{varying continuation seed}    & & \\
      \;\;+ NoiseRater ($p{=}4$),          & \multirow{2}{*}{$60.24 \pm 0.03$} & \multirow{2}{*}{$-1.43$} \\
      \;\;\;\;\textit{varying rater seed}           & & \\
      \midrule
      Min-SNR                          & $59.08 \pm 0.08$ & \multirow{2}{*}{$-1.22$} \\
      \;\;+ NoiseRater ($p{=}4$)       & $\mathbf{57.86 \pm 0.11}$ & \\
      \midrule
      P2                               & $57.48 \pm 0.12$ & \multirow{2}{*}{$-1.27$} \\
      \;\;+ NoiseRater ($p{=}4$)       & $\mathbf{56.21 \pm 0.09}$ & \\
      \midrule
      Logit-normal                     & $57.23 \pm 0.06$ & \multirow{2}{*}{$-1.21$} \\
      \;\;+ NoiseRater ($p{=}4$)       & $\mathbf{56.02 \pm 0.04}$ & \\
      \bottomrule
  \end{tabular}
\end{table}

\begin{table}[h!]
  
  \centering
  \small
  \caption{{Repeatability at equal compute. FID-50K at 17.6 GPU-h for DiT-S/2 on ImageNet $256{\times}256$ with the $p{=}4$ rater, under the protocol of Table~\ref{tab:tiny_compute_matched}, reported as mean $\pm$ standard deviation over three independent runs (two for Logit-normal + NoiseRater).}}
  \label{tab:seeds_equal_compute}
  \begin{tabular}{l c c c}
      \toprule
      & \multicolumn{2}{c}{ImageNet $256{\times}256$, FID-50K $\downarrow$ at 17.6 GPU-h} & \\
      \cmidrule(lr){2-3}
      \textbf{Method} & \textbf{Baseline} & \textbf{+ NoiseRater} & $\Delta$ \\
      \midrule
      Vanilla        & $61.67 \pm 0.02$ & $\mathbf{60.48 \pm 0.10}$ & $-1.19$ \\
      Min-SNR        & $59.08 \pm 0.08$ & $\mathbf{58.22 \pm 0.11}$ & $-0.86$ \\
      P2             & $57.48 \pm 0.12$ & $\mathbf{56.65 \pm 0.08}$ & $-0.83$ \\
      Logit-normal   & $57.23 \pm 0.06$ & $\mathbf{56.38 \pm 0.08}$ & $-0.85$ \\
      \bottomrule
  \end{tabular}
\end{table}

\clearpage

\section[Fidelity and diversity under noise selection]{{Fidelity and diversity under noise selection}}
\label{app:diversity}

Top-1 selection changes the distribution of the training noise (Equ.~\ref{eq:stage2_objective}), and FID alone does not show whether the gain comes at the cost of sample diversity. We therefore evaluate the models of Table~\ref{tab:strategy_comparison} and Table~\ref{tab:effect_of_k_compute} at +80k with precision and recall, density and coverage, and the Inception Score (IS), using 50k generated and 50k real images. We also include a soft-weighting variant, in which the frozen rater weights all $k{=}4$ candidates with the group-wise weights of Section~\ref{sec:noise_weighting} instead of keeping only the top-scored one.

\begin{table}[h!]
  
  \centering
  \small
  \caption{{Fidelity and diversity metrics at +80k for DiT-S/2 on ImageNet $256{\times}256$ (class-conditional without CFG). All rater variants use the full-size rater.}}
  \label{tab:diversity}
  \begin{tabular}{l cccccc}
      \toprule
      \textbf{Method} & \textbf{FID} $\downarrow$ & \textbf{Precision} $\uparrow$ & \textbf{Recall} $\uparrow$ & \textbf{Density} $\uparrow$ & \textbf{Coverage} $\uparrow$ & \textbf{IS} $\uparrow$ \\
      \midrule
      Vanilla                         & 61.67 & 0.725 & \textbf{0.232} & 1.062 & \textbf{0.459} & 19.63 \\
      \midrule
      Rater (top), $k{=}2$            & 60.38 & 0.734 & 0.224 & 1.094 & 0.455 & 19.60 \\
      Rater (top), $k{=}4$            & 59.97 & 0.741 & 0.207 & 1.152 & 0.438 & 19.08 \\
      Rater (top), $k{=}8$            & 60.97 & \textbf{0.753} & 0.191 & \textbf{1.269} & 0.411 & 17.92 \\
      \midrule
      Rater (bottom), $k{=}4$         & 64.43 & 0.720 & 0.226 & 1.040 & 0.446 & 18.86 \\
      Rater (soft weighting), $k{=}4$ & \textbf{59.42} & 0.739 & 0.221 & 1.119 & 0.453 & \textbf{19.71} \\
      \bottomrule
  \end{tabular}
\end{table}

\paragraph{The ranking is not a fidelity--diversity dial.}
As shown in Tab.~\ref{tab:diversity}, Rater (bottom) is below Vanilla on every metric, including both precision and recall. If the rater's score only moved the model along a trade-off between fidelity and diversity, its bottom-scored noise would lower precision and raise recall. Low-scored noise is instead less useful for training on both axes, which supports the finding of Section~\ref{sec:experiment_unequal}.

\paragraph{Top-1 selection trades diversity for fidelity.}
With top-1 selection, precision and density increase with $k$, while recall and coverage decrease. At $k{=}4$, precision rises from $0.725$ to $0.741$ and recall falls from $0.232$ to $0.207$. At $k{=}8$, recall and coverage decline further and FID is worse than at $k{=}4$. This is consistent with the hypothesis in Section~\ref{sec:experiment_improve} that selecting from a large pool narrows the training noise distribution too far. The pool size $k$ therefore controls this trade-off, and the loss of recall and coverage is a limitation of hard top-1 selection.

\paragraph{Soft weighting.}
Weighting all four candidates gives the best FID ($59.42$) and IS, with recall ($0.221$) and coverage ($0.453$) close to Vanilla ($0.232$ and $0.459$). This variant requires a forward and backward pass of the diffusion model for every candidate, so it is more expensive per step than top-1 selection, which remains our default.

\section[Denoising error under standard Gaussian noise]{{Denoising error under standard Gaussian noise}}
\label{app:denoising_error}

Top-1 selection trains the diffusion model on noise drawn from $q_\eta$ (Equ.~\ref{eq:stage2_objective}) rather than from $\mathcal{N}(0, I)$, while sampling still assumes the standard objective of Equ.~\ref{equ:diff_loss}. The two objectives are not guaranteed to share a minimizer. This section bounds the shift of the noise distribution and measures its effect on the standard objective.

\paragraph{Bound on the selected-noise distribution.}
The selected noise is one of $K'$ candidates drawn i.i.d.\ from $\mathcal{N}(0, I)$ at every step. For any measurable set $A$, the selected noise lies in $A$ only if at least one candidate does, so by the union bound
\begin{align}
q_\eta(A \mid \mathbf{x}_0, t, \tilde{c}) \;\le\; K' \cdot \mathbb{P}_{\epsilon \sim \mathcal{N}(0, I)}\big[\epsilon \in A\big].
\label{eq:density_bound}
\end{align}
The density of $q_\eta$ is therefore at most $K'$ times the standard Gaussian density, and $q_\eta$ is absolutely continuous with respect to $\mathcal{N}(0, I)$. This bound limits how far the training noise can move. We further added experiments to measure how much this shift changes the denoising error under the standard objective.

\paragraph{Setup.}
We evaluate the standard objective, i.e., the $\epsilon$-prediction error under noise drawn from $\mathcal{N}(0, I)$ without any selection, on images that were not used for training the diffusion model or the rater. We use the 50{,}000 images of the ImageNet validation set, and the 10{,}000 images of ImageNet-V2~\citep{recht2019imagenet} as a related dataset collected independently of ImageNet. For each image, we draw one timestep uniformly and one Gaussian noise, and all models are evaluated on identical (image, timestep, noise) triples. We compare the 400k checkpoint from which both continuations start, the vanilla continuation, and the \emph{NoiseRater} continuation (top-1 selection with $k{=}4$), both at +80k steps.

\paragraph{Results.}
As shown in Tab.~\ref{tab:denoising_error}, both continuations reduce the error of the starting checkpoint, by $0.20\%$ for vanilla training and $0.11\%$ for \emph{NoiseRater} on the ImageNet validation set. The error of the \emph{NoiseRater} continuation is thus about $0.09\%$ higher than that of the vanilla continuation, and the same holds on ImageNet-V2. Training on selected noise leads to a small increase in the error under standard Gaussian noise at the tested checkpoints. The comparison is made at equal continuation steps; a comparison at equal GPU-hours is left open. Under the unchanged sampler, the \emph{NoiseRater} continuation still has lower FID (Appx.~\ref{app:seeds}), while its recall and coverage are reduced (Appx.~\ref{app:diversity}).

\begin{table}[h!]
  
  \centering
  \small
  \caption{{Denoising error under standard Gaussian noise on unseen images, for DiT-S/2 trained on ImageNet $256{\times}256$. We report the $\epsilon$-prediction MSE ($\times 10^{-2}$) averaged over all timesteps, and its change relative to the starting checkpoint; lower is better.}}
  \label{tab:denoising_error}
  \begin{tabular}{l cc cc}
      \toprule
      & \multicolumn{2}{c}{\textbf{ImageNet val (50k)}} & \multicolumn{2}{c}{\textbf{ImageNet-V2 (10k)}} \\
      \cmidrule(lr){2-3} \cmidrule(lr){4-5}
      \textbf{Model} & MSE $\downarrow$ & Change & MSE $\downarrow$ & Change \\
      \midrule
      Starting checkpoint (400k)        & 14.332 & --        & 14.620 & -- \\
      Vanilla, +80k                     & \textbf{14.303} & $-0.20\%$ & \textbf{14.592} & $-0.19\%$ \\
      NoiseRater ($k{=}4$), +80k        & 14.316 & $-0.11\%$ & 14.605 & $-0.10\%$ \\
      \bottomrule
  \end{tabular}
\end{table}

\end{document}